\documentclass[10pt,twocolumn,letterpaper]{article}

\usepackage[pagenumbers]{cvpr} % To force page numbers, e.g. for an arXiv version

\usepackage{graphicx}
\usepackage{amsmath}
\usepackage{amssymb}
\usepackage{booktabs}
\usepackage{mathtools}
\usepackage{algorithm}
\usepackage{algorithmic}
\usepackage{float}

\usepackage{placeins}
\usepackage{amsfonts}
\usepackage{epsfig}
\usepackage{color}
\usepackage{xcolor}
\usepackage{stmaryrd}
\usepackage{multirow}
\usepackage{dsfont}
\usepackage{standalone}

\makeatletter
\@namedef{ver@everyshi.sty}{}
\makeatother
\usepackage{tikz}
\usepackage{changes}

\definechangesauthor[name={Georg Pichler}, color=violet]{GP}

\usepackage{pgfplots}[]
\usepackage{pgfplotstable}

\usepgfplotslibrary{fillbetween}
\pgfplotsset{width=0.8\textwidth,compat=newest}
\pgfplotstableset{col sep=comma}

\pgfplotsset{
    basicAxis/.style={
    major grid style={line width=.2pt,draw=gray!50},
    grid=both,
    width = \linewidth,
    height = .6\linewidth,
    x label style ={yshift=-\baselineskip, anchor=south},
    y label style = {yshift=0.0cm, anchor=south},
    legend style={nodes={scale=0.7, transform shape}, at={(0.50,0.40)},anchor=north west},
    },
    basicTable/.style={
    each nth point=50,
    filter discard warning=false,
    unbounded coords=discard
    }
}

\newcommand{\defeq}{\stackrel{\rm def}{=}}
\newcommand{\EE}[1]{\mathbb{E}\left\{ #1 \right\}}

\newcommand{\G}{{\cal G}}
\newcommand{\X}{{\cal X}}

\newcommand{\rset}{\mathbb{R}}

\newcommand{\MLattacker}{ML Attacker}

\def\TrainSet{\mathcal{D}_n}
\def\targetModel{\widehat{g}_\theta}
\def\testSample{(x_\mathrm{test},y_\mathrm{test})}
\def\AdvStrat{\psi}
\def\XXX{\mathcal X}
\def\SSS{\mathcal S}
\def\YYY{\mathcal Y}

\def\argmax{\mathop{\rm arg\, max}}

\newcommand{\un}{\mathds{1}}

\usepackage[pagebackref,breaklinks,colorlinks]{hyperref}

\usetikzlibrary{arrows.meta,
                calc, chains,
                quotes,
                positioning,
                shapes.geometric}

\usepackage[capitalize]{cleveref}
\crefname{section}{Sec.}{Secs.}
\Crefname{section}{Section}{Sections}
\Crefname{table}{Table}{Tables}
\crefname{table}{Tab.}{Tabs.}

\def\cvprPaperID{6582} % *** Enter the CVPR Paper ID here
\def\confName{CVPR}
\def\confYear{2022}

\begin{document}

\title{Leveraging Adversarial Examples to Quantify\\  Membership Information Leakage\\
{\small In proceedings of CVPR 2022}
}

\author{Ganesh Del Grosso\,$^1$\\
%Inria, Ecole Polytechnique\\
%1 Rue Honoré d'Estienne d'Orves\\
%91120 Palaiseau, France\\
%{\tt\small ganesh.del-grosso-guzman@inria.fr}
\and 
Hamid Jalalzai\,$^1$\\
%Inria\\
%1 Rue Honoré d'Estienne d'Orves\\
%91120 Palaiseau, France\\
%{\tt\small hamid.jalalzai@inria.fr }
\and
Georg Pichler\,$^2$\\
%TU Wien\\
%Gusshausstrasse 30\\
%1040 Wien, Austria\\
%{\tt\small georg.pichler@ieee.org}
\and 
Catuscia Palamidessi\,$^1$\\
%Inria, Ecole Polytechnique\\
%1 Rue Honoré d'Estienne d'Orves\\
%91120 Palaiseau, France\\
%{\tt\small catuscia@lix.polytechnique.fr}
\and 
Pablo Piantanida\,$^3$\\
%International Laboratory on Learning Systems (ILLS) \\
%McGill  ETS  MILA  CNRS  Université Paris-Saclay \\
%3 Rue Joliot Curie\\
%H3C 1K3 Quebec, Canada\\
%{\tt\small piantani@mila.quebec}
}
\author{ Ganesh Del Grosso$\,^{1,}$\thanks{Equal contribution.} \quad Hamid Jalalzai$\,^{1,}$\footnotemark[1]   \quad Georg Pichler$\,^{2,}$\footnotemark[1] \quad Catuscia Palamidessi$\,^{1}$  \quad Pablo Piantanida$\,^{3}$ \\
$^{1}$ Inria - Laboratoire d'Informatique de l'\'Ecole polytechnique, France \\
$^{2}$ TU Wien, %Gusshausstrasse 30
%1040 Wien, 
Austria
\\
$^{3}$ International Laboratory on Learning Systems (ILLS) \\ McGill - ETS - MILA -  CNRS - Université Paris-Saclay - CentraleSupélec,
%3 Rue Joliot Curie, H3C 1K3 Quebec, 
Canada
\\
{\tt\small $^{1}$\{ganesh.del-grosso-guzman, hamid.jalalzai, catuscia.palamidessi\}@inria.fr} \\
{\tt\small $^{2}$georg.pichler@ieee.org} \quad {\tt\small $^{3}$piantani@mila.quebec} 
}

\maketitle

\begin{abstract}
The use of personal data for training machine learning systems comes with a privacy threat and measuring the level of privacy of a model is one of the major challenges in machine learning today. Identifying training data based on a trained model is a standard way of measuring the privacy risks induced by the model. We develop a novel approach to address the problem of membership inference in pattern recognition models, relying on information provided by adversarial examples. The strategy we propose consists of measuring the magnitude of a perturbation necessary to build an adversarial example. Indeed, we argue that this quantity reflects the likelihood of belonging to the training data. Extensive numerical experiments on multivariate data and an array of state-of-the-art target models show that our method performs comparable or even outperforms state-of-the-art strategies, but without requiring any additional training samples.

%Extensive numerical experiments on multivariate data and an array of state-of-the-art target models show that our method outperforms strategies previously considered in the literature.  Furthermore, while most competing strategies rely on training data to build attack models, we show that comparable performance can be achieved  using  similar  techniques,  but  without  additional training data
\end{abstract}

\section{Introduction}
\label{sec:introduction}
With the deluge of data and increase of computational power within the last decades, performance of modern machine learning shows dramatic improvement in a wide range of applications such as computer vision \cite{forsyth2011computer, redmon2016you, Jegou_2017_CVPR_Workshops} and natural language processing \cite{bengio2003neural, devlin2018bert, yang2019xlnet}. Along with this improvements new methods emerge, leading to remarkable change in societal applications ranging from industry \cite{candes2006stable, karpathy2014large} to modern medicine \cite{lustig2008compressed,wong2018machine, jumper2021highly} to art \cite{huang2017arbitrary, gatys2016image}, all of which may be considered \emph{sensitive} domains, given the nature of the data.

%Pattern recognition lies among the most encountered frameworks of machine learning \cite{friedman2001elements}. 
% As further detailed in Section~\ref{subsec:patternreco}, its purpose is to predict a discrete label $Y$ given some multivariate input $X$ with lowest error based on observing $\mathcal{D}_{train}$ composed of $n >0$ i.i.d copies of the random pair $Z = (X, Y)$.  
While the benefits of machine learning are set upfront, other societal aspects such as fairness \cite{vogel2020learning, mehrabi2021survey} or safety \cite{gebru} should not be trampled on \cite{amodei2016concrete}. {It is common consensus} that models require vast amounts of training data \cite{deng2009imagenet, abu2016youtube} to reach {state-of-the-art} performance; meanwhile they do not necessarily guarantee the anonymity of the data provider~\cite{hern2017royal}. This represents a serious privacy issue and controlling such leakage of information with modern regulation presents a new challenge \cite{mcgraw2021privacy, harding2019understanding}. With recent data protection regulations \cite{regulation2016regulation,CCPA} personal data are required to be protected while being used by machine learning models \cite{tankard2016gdpr}. To improve safety in machine learning, the study of attack strategies that exploit models in order to infer training data, or even corrupt them, has become an active area of research. 
% Adversarial examples \cite{szegedy2013intriguing, ben} may. 

In this paper we investigate membership inference attacks (MIAs) \cite{shokri2017membership, shokri2015privacy, salem2018ml, liu2019socinf, Song2021, song2019privacy, long2017towards, truex2019demystifying, long2018understanding, yeom2018privacy}, in which an attacker tries to determine whether or not a sample was  part of the training set of a target model \cite{shokri2017membership}. %The importance of understanding the robustness to MIAs is to protect machine learning models from  more severe privacy violations by preventing malicious adversaries from linking potential private information present in the training set to a particular data sample \cite{Li2013}.
By leveraging adversarial attacks, we propose an MIA strategy that achieves similar performance to the {state-of-the-art}, but without using training samples to construct the attack. We only require accessing  the target model and the testing sample. 

\begin{figure*}[!h]
\centering
\includegraphics[width=.9\textwidth]{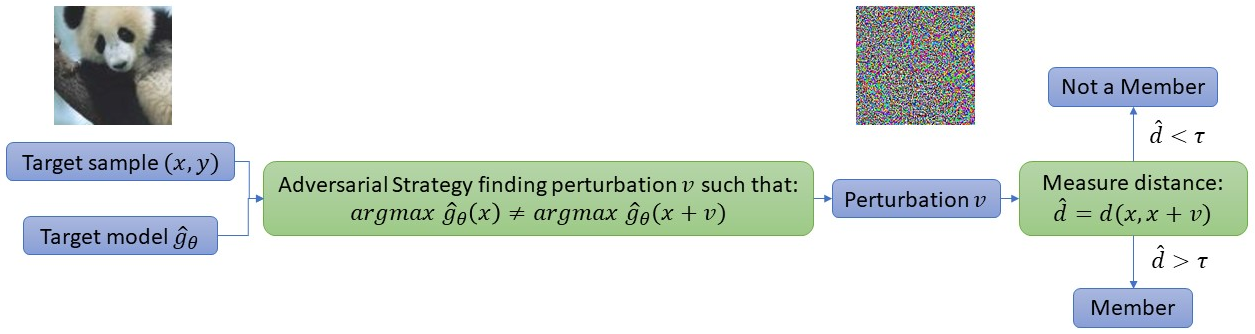}
\caption{Illustrative diagram of our method to perform membership inference attack based on adversarial perturbations.}
  \label{fig:diagram}
\end{figure*}

Adversarial attacks maximize the loss function of a model with respect to an input sample, in order to find a perturbation that changes the class predicted by the model. Interestingly, we empirically observed that changing the predicted class requires a larger perturbation for samples that are part of the training set since the model was tuned to minimize the empirical loss function computed using these samples. Hence the idea is to measure this perturbation, 
i.e., the distance between an adversarial example and its original counterpart, and test whether it is lager than a certain threshold. We call this measure \emph{Adversarial Distance}.  Figure~\ref{fig:diagram} provides an overview of our strategy\footnote{The illustrations for the pipeline's input and adversarial noise are provided by \cite{goodfellow2014explaining}. The noise illustrated in Figure~\ref{fig:diagram} is obtained with a fast adversarial example against GoogLeNet’s classification algorithm. The added noise changes the classifier's output from class ``\textit{panda}" to class ``\textit{gibbon}".}. 

In contrast to other recent works~\cite{shokri2017membership,NasrShokri, Rezaei}, which provide the attacker with a subset of the training set of the target model, our approach does not require any training data. Intuitively, if a model is susceptible to MIAs without resorting to training resources, we would expect it to be even more vulnerable in presence of additional data. As a matter of fact, we will show that in many cases additional samples are not necessary in order to accurately determine the membership of target samples. 
% Empirically, we show that in some settings, expanding on ideas from previous works, similar performance can be obtained without giving labeled data to the attacker.

\subsection{Contributions}
Below we list the contributions of this work:
\begin{itemize}
    \item We propose a novel MIA (\cref{sec:MembershipInferenceAttacksfromAdversarialExamples}) that performs consistently well regardless of the architecture of the target model, and does not require training samples. This strategy exploits the distance between adversarial examples and the corresponding raw inputs. 
    %To the best of our knowledge, this is the first work introducing this idea to perform MIAs. 
    % A detailed explanation is provided in Section~\ref{sec:AdvDist}. 
    % \georg{that is not a contribution. we need to phrase differently.}
    % \hamid{The point is to say that the adversarial distance was not used for MIA before this work.}
   
    \item We perform a thorough revision of MIA strategies previously proposed in the literature and evaluate their performance (\cref{sec:experiments}). Through this evaluation we show that several well-known machine learning models are  vulnerable to MIAs.
    %The numerical results of this evaluation are relegated to.
 
    \item Empirically, we show that in most of the investigated scenarios the proposed MIA outperforms, or it is at least competing with, state-of-the-art methods that rely on a large amount of data samples to perform the attack. On the other hand, for large models (\textit{e.g.}, DenseNet), we observe that  training samples can grant a significant advantage to the attacker.
\end{itemize}

% \deleted{with overwhelming probability. Following quantities introduced in the paper, we recall that $\tau \in \rset_+$}.}. 

% \emph{(i)} that our attack strategy is able to extract membership information consistently against all target models, even without having access to additional samples and \emph{(ii)} that certain attack models can achieve nearly perfect accuracy against large neural networks. These results are presented in \cref{sec:experiments}.
% While recent works suggest that MIA against machine learning models are ineffective, even when the attacker has access to a large amount of additional (In/Out)-training samples \cite{Rezaei}, we argue the contrary. For this purpose 

% \textcolor{red}{\#I THINK ALL CONTRIBUTION SUBSECTION SHOULD BE COMPLETELY REWRITTEN in a short way. At the same time, the sections where the main contributions are should be indicated. \#}

% \paragraph{Notations} The following notations are used throughout the paper.

\subsection{Related Work}

Along the review cycle, the reviewers brought to our attention the work from the authors of \cite{LabelOnlyLi2021} where they propose a similar strategy for MIAs to the one depicted in this article. We thank the reviewers for pointing out such concurrent work. However, their work differs from ours in two crucial ways: First, our attack exploits white-box access to the target model, which allows for different, more powerful adversarial strategies. Namely, they use HopSkipJump \cite{HopSkipJump}
and QEBA \cite{QEBA}, while we use Auto-PGD \cite{AutoAttack}. Second, we evaluate the performance of our strategy on real-world models, testing our attack strategies and those of our competitors on pre-trained state-of-the-art ML models for image classification (AlexNet, DenseNet, ResNet, ResNext), while their work uses its own target model, which makes it difficult to compare the results directly. 

\noindent\textbf{Membership Inference Attacks.} The work~\cite{shokri2017membership} introduced MIAs against machine learning models. The attacks proposed in~\cite{shokri2017membership} consist of training an attack model that observes the input-output relation of the target model. In this black-box scenario, the attacker can access the target model only by querying it. In order to be trained, the attacker requires a part of the training set of the target model. When this is not available, the attacker resorts to training its own \textit{shadow models}, which share the same architecture as the target model, but provide the attacker with full knowledge of their training set. This seminal work was further extended by \cite{NasrShokri}, which considers an attacker with white-box access to the target model,  using intermediate outputs of the target model and gradients of the loss function as input to the attacker.

A more recent contribution~\cite{Song2021} proposes the use of entropy and ``Modified Entropy''  in MIAs, while \cite{Rezaei} studies the use of gradients with respect to the input samples, gradients with respect to model parameters, intermediate outputs and ``distance to the decision boundary'' in MIAs. Note that our Adversarial Distance strategy does not attempt to estimate the distance to the decision boundary, but to estimate the magnitude of the perturbation required to produce an adversarial example, \textit{i.e.}\ a sample that is classified incorrectly with very high confidence. Remarkably, while \cite{Rezaei} suggests that MIAs are ineffective against machine learning models, 
% even when the attacker has access to a large amount of additional %(In/Out)-training samples.
our revision of state-of-the-art MIA strategies provides evidence to the contrary. It is important to note that we repeated some of the same experiments of~\cite{Rezaei}, obtaining different results.

The aforementioned works are taken as baselines for assessing the performance of our method. We reproduce their results and, like those works, we use the output of the target model, loss value, norm of the gradient and modified entropy. However, we directly implement MIAs as binary decision tests, without training attack models.

The authors of~\cite{ShokriUnleashed} propose to exploit the model's predictions on adversarial examples to perform MIAs. Their method consists of using the predictions on  %targeted (and untargeted) 
adversarial examples to distinguish members from non-members of the training set. This strategy is found to be  significantly more effective against models that are robust against adversarial attacks. In our work we take a different approach, by measuring the size of the perturbation necessary to produce an adversarial example, rather than looking at the model's prediction on an adversarial example. In contrast to \cite{ShokriUnleashed}, our aim is to develop effective MIAs, and not to compare the vulnerability of different models.

In \cite{shafran2021membership} the reader can find an extensive discussion on the success and complexity of MIAs based on the difficulty of the underlying machine learning task. In \cite{truex2019demystifying} a strategy similar to \cite{shokri2017membership} is used, but the emphasis is on investigating how the vulnerability of the model is influenced by the model choice and dataset, rather than providing novel attack strategies. The work \cite{PrivacyYeom} studies the connection between MIAs, attribute inference, differential privacy and overfitting, and proposes an MIA method that uses the value of the loss function to distinguish members from non-members of the training set. We use a similar strategy, but without requiring the average value of the loss of the target model over its training set. Finally, a comprehensive study of MIAs against GANs and other generative models is provided by \cite{GANleaksChen}. Although similar attack strategies can be used in this context, generative models are beyond the scope of the present work.
    
\noindent\textbf{Adversarial Attacks.} Adversarial examples were first introduced in \cite{szegedy2013intriguing}, showing that most ML models, and more specifically neural networks, are vulnerable to minor changes to their inputs~\cite{goodfellow2014explaining}. Since then, a myriad of works has emerged in this field \cite{Carmon2019UnlabeledDI,Moosavi-Dezfooli_2016_CVPR,NEURIPS2020_WU,pmlr-v119-rice20a} and opened the path for a better understanding of broader issues in generative models and mechanisms to fool algorithms, including deepfake technologies \cite{westerlund2019emergence, khachaturov2021markpainting}. We will use the adversarial attack proposed by \cite{AutoAttack}, as it is highly adaptable and does not require fine-tuning of additional parameters. 

% \textcolor{red}{\#I Abstract and introduction should be shorten to two pages (even less would be better). Remember that we have 8 pages at all!! \#}

%Several contributions such as  led the path for better consideration to safety in machine learning as illustrated by deepfake \cite{westerlund2019emergence, khachaturov2021markpainting} or example2 \cite{} \hamid{TODO: find a second example and cite work.}

%\paragraph{Outline} Section~\ref{sec:preliminaries} formally introduces membership inference attacks and adversarial attacks. In Section~\ref{sec:MembershipInferenceAttacksfromAdversarialExamples} we present our main contributions along with the specific links between membership inference and adversarial examples. We detail and perform relevant numerical experiments in Section~\ref{sec:experirements} to highlight the performance of our method. 

%Finally, we provide future directions of research and conclusion in Section~\ref{sec:conclusion}. 
% Relevant material may be found in 

\section{Definitions and Preliminaries}
\label{sec:preliminaries}
In this section, we introduce the framework and notation used throughout the rest of the paper and the formal definitions for membership inference and adversarial attacks.

\subsection{Pattern Recognition}
\label{subsec:patternreco}
Among the classical frameworks of machine learning, pattern recognition \cite{devroye2013probabilistic} consists of predicting a discrete random variable $Y \in \mathcal{Y}$ based on observing some multivariate random variable $X \in \X$, using
% where $\X \subset \rset^p$ with $p > 1$. 
%$X$ hopefully contains enough valuable information concerning $Y$ for 
a classifier $g_\theta$ from a class of classifiers $\G$,
% $=\left\{ g_\theta\, | \, \theta \in \Theta\right\}$
parameterized by $\theta$. The classifier outputs a probability distribution on the label set $\mathcal{Y}$ (\textit{i.e.} for any $g_\theta \in \G$, $g_\theta: \X \mapsto \mathcal P(\mathcal{Y})$) to predict $Y$ with low classification risk $R(g_\theta) = \EE{\ell(g_\theta(X), Y)}$, where loss function $\ell$ measures the error occurring between the true label $Y$ and the one provided by the classification rule $g_\theta$.
% (\textit{e.g.} one may consider the well-known $0/1$ loss defined as $\ell(Y, g(X))= \un\{Y \not= g(X)\}$). 
As the joint distribution $P_{XY}$ is unknown in practice, in order to select a possible classifier, one relies on $\TrainSet = \{(x_i, y_i) \}_{i=1}^n$ composed of $n \geq 1$ realizations of $(X, Y)$. The empirical risk minimization paradigm \cite{vapnik1992principles} suggests to select the classifier minimizing the empirical risk $\widehat{R}(g_\theta)= \frac{1}{n}\sum_{i=1}^n \ell(g_\theta(x_i), y_i)$, among all the possible classifiers in $\G$. Let $\targetModel$ denote the classifier minimizing $\widehat{R}$ for a training set $\TrainSet$.

\subsection{Membership Inference Attacks}

% In the black-box scenario, the attacker is only allowed to query the model, and has access to its output given a certain input. In the white-box scenario the attacker has access to the model architecture and parameters, allowing the computation of gradients and intermediate outputs of the model. 

Membership inference attacks  (MIAs) can be used to measure the privacy leakage of ML models \cite{shokri2017membership}. The goal of a MIA is to determine whether or not a sample (or group of samples) belongs to the training set of target model. Formally, MIAs can be stated as a binary decision test. Given a test sample $\testSample$, and the target model $\targetModel$ defined above, the goal of the attacker is to determine if $\testSample\in\TrainSet$. Let $\varphi$ be a scoring criteria, which takes as input the target model, the test sample and outputs a prediction score. This prediction score can be compared to a threshold $\tau\in\rset_{+}$ to predict if the test sample belongs to the training set of the target model. Formally,
\begin{equation}
\begin{aligned}
      &\mathrm{if}\;\;\; \varphi(\testSample,\targetModel)\geq\tau \;\;\;\mathrm{then}\;\;\; \testSample\in\TrainSet
      \\
   % \nonumber\\
  &\mathrm{otherwise}\;\;\; \testSample\notin\TrainSet\; .
    \label{eq:MIA}
\end{aligned}
\end{equation}
% The threshold $\tau$ is a hyper-parameter of our approach. 
The hyper parameter $\tau$ of our approach selects the operating point in ROC curve. In practice, we make our analysis independent of $\tau$ by comparing performance for the whole range of possible $\tau$ values. 

%gets too small, a bias is induced by  observations which do not necessarily belong to the train set. On the other hand, too large values lead to misclassification of the positive class.

%dedicated to explaining some basic scoring criteria that summarize the observations made in previous works.

Hereafter, we consider different scoring criteria.

\noindent \textbf{Softmax Response.} Our main claim is that models tend to give more confident predictions over samples that belongs to their training set. This strategy aims to exploit the confidence of the predictions to identify members of the training set of the target model: 
\begin{align}
    \varphi(\testSample,\targetModel) = \max_{i\in \YYY } \targetModel^i(x_\mathrm{test})\;,
\end{align}
where $\targetModel^i$ is the $i$-th component of the output of the model parametrized by $\theta$. This observation has previously been used to build MIAs in  \cite{shokri2017membership,NasrShokri}, training an attack model, while \cite{ShokriUnleashed} directly compares the score to a threshold.

\noindent \textbf{Modified Entropy.} An alternative idea is to look at the uncertainty of the model. Intuitively, this should be lower for samples that were present in the training set. \cite{Song2021} proposes a metric called modified entropy, which decreases with the prediction probability of the correct class and increases with the prediction probability of any other class:
\begin{align}
    \varphi(\testSample,\targetModel) & = -\left(1-\targetModel^{y_\mathrm{test}}(x_\mathrm{test})\right)\log\left(\targetModel^{y_\mathrm{test}}(x_\mathrm{test})\right)\nonumber\\
    &-\sum_{i\neq y_\mathrm{test}}\targetModel^{i}(x_\mathrm{test})\log\left(1-\targetModel^{i}(x_\mathrm{test})\right).
    \label{eq:score:mentr}
\end{align}
Unlike other metrics, modified entropy \eqref{eq:score:mentr}  takes into account whether the target model is predicting the correct class.

\noindent\textbf{Loss.} The learning objective of ML models is to minimize a loss function,
\begin{align}
    \varphi(\testSample,\targetModel) = -\ell\left(y_\mathrm{test},\targetModel(x_\mathrm{test})\right)\;,
    \label{eq:score:loss}
\end{align}
 over samples from the training set. Hence, we expect the value of the loss to be lower for samples in the training set. The minus sign in front of the loss is added to make this definition consistent with \cref{eq:MIA}.  An attack proposed in \cite{PrivacyYeom} compares the loss on the test sample to the average loss on the training set. The idea was also exploited in \cite{shokri2017membership}.

\noindent\textbf{Gradient Norm.} The loss function is minimized via Stochastic Gradient Descent, or similar iterative optimization algorithms. Around the optimal points, the gradient of the loss function with respect to its model parameters should approach $0$. This attack strategy measures the $\ell_2$ norm of the gradient of the loss function w.r.t. to the model parameters over different samples and expects this norm to be smaller for members of the training set,
\begin{align}
    \varphi(\testSample,\targetModel) = -\Vert\nabla_{\theta}\ell\left(y_\mathrm{test},\targetModel(x_\mathrm{test})\right)\Vert^2_2 \;.
\end{align}
Since we expect the norm of the gradient to be smaller for members of the training set, the minus sign is added to make this definition consistent with \cref{eq:MIA}. This observation was first used in \cite{shokri2017membership} as part of their MIA. 

Although these ideas are not novel, most of them have not been used to make a binary decision test. Our aim is to assess and compare in a systematic way the power of these observations and whether or not it is possible to perform MIAs with them without requiring a training set for the attacker.

% in this we use a metric that is independent of the threshold, the area under receiver operating characteristic curve (AUROC), to measure their performance. 

\subsection{Adversarial Examples}

The framework of untargeted adversarial examples can be set as follows: Given an input $x \in \mathcal{X}$ and target model $\targetModel: \mathcal{X} \mapsto \mathcal{P}(\mathcal{Y})$, the goal of adversarial strategy $\AdvStrat_{p,\epsilon}$ is to produce some perturbation $v\in \mathcal{X}$ such that the prediction provided by $\targetModel(x + v)$ changes from that provided by $\targetModel(x)$. Additionally, we require that the target model is confident on its prediction of the adversarial example. Formally, we define an untargeted adversarial strategy for a classifier $g \in \mathcal{G}$ as a function $\AdvStrat_{p,\epsilon}\colon \mathcal{X} \to \mathcal{X}$
 on the input space $\mathcal{X}$, such that for any $x \in \mathcal{X}$ it obtains $ v\defeq \AdvStrat_{p,\epsilon}(x) \in \mathcal{X}$ with
\begin{align}
   \argmax_{i\in \YYY } g^i(x + v) &\not= \argmax_{i\in \YYY } g^i(x) \text{ and} \\
   \Vert v \Vert_p &\le \epsilon \;,
 \end{align}
\textit{i.e.}, the constrained perturbation $v$ changes the prediction of the target model to a wrong class.

\iffalse
\textcolor{red}{\Cref{fig:illustration} illustrates various adversarial examples built from a training sample in $\rset^2$. The background color represents the output of a binary neural network classifier trained on the scattered training data. \# This does not add much if the testing samples (out of the training set) do not paper as well. Our paper is not about building adversarial samples. \#}
\begin{figure}
  %\centering
  \includegraphics[width=0.49\textwidth]{./figures/illustration.pdf}
  \caption{Illustration of bivariate data and output of a binary classifier (background color) with adversarial examples built from a training sample.}
  \label{fig:illustration} 
  \textcolor{red}{\# Why not having the testing samples to see if the distance is larger ? Otherwise, I do not see the value of this picture since people know what is a decision boundary. \#}
\end{figure}
\fi

Adversarial examples are computed constraining $\Vert v\Vert_p<\epsilon$, with $\epsilon\in\rset_+$ and the $\ell_p$ norm $\Vert\cdot\Vert_p$ (see \cite{akhtar2018threat} for an extensive review on adversarial strategies). The purpose of this constraint is twofold: to perturb the original image in a way that is imperceptible for the human eye and to control the power of the attacker. In our case, the goal is not to produce subtle perturbations, the adversarial examples may be significantly different from their original counterparts. Indeed, our goal is to observe the size of the perturbation necessary to force the target model to drastically change its prediction, and use it as a criteria to distinguish members from non-members of the training set. Since $\AdvStrat_{p,\epsilon}$ will tend to compute the smallest perturbation possible such that $\targetModel$ changes its prediction, arbitrarily high $\epsilon$ can be allowed while still observing a significant difference between the size of the perturbation of samples in and outside the training set.

% The framework of untargeted adversarial example can be set as follows, given an input $x \in \mathcal{X}$ and a classifier with minimum generalization error $g : \mathcal{X} \mapsto \mathcal{Y}$, for some noise/perturbation $v \in \mathcal{X}$ provided by an adversarial strategy $\mathcal{V}$, the prediction provided by $g(x + v)$ may be different from $g(x)$.

% Further constraints can be added on the adversarial strategy $\mathcal{V}$ from Definition~\ref{def:adv_strat} in order to minimize the discrepancy between $x$ and $x + v$.

For our experiments we use \textit{Auto-Attack} to build adversarial examples\footnote{Code available at \url{https://github.com/fra31/auto-attack}.} \cite{AutoAttack}. The Auto-Attack library offers an ensemble of different strategies to compute adversarial examples. Particularly, we use auto Projected Gradient Descent (Auto-PGD). Given an objective function for the adversary $f_a:\XXX\mapsto\rset$ and a constraint in the form $\SSS\subset\XXX$, Auto-PGD  iteratively solves $\max_{x\in\SSS}f_a(x)$ by applying $x^{(k+1)}=P_{\SSS}\left(x^{(k)}+\eta^{(k)}\nabla_{x^{(k)}}f_a(x^{(k)})\right)$, for $k=[1,\dots,N_\mathrm{iter}]$, where $P_\SSS$ is the projection onto the surface of $\SSS$, and typically $f_a(x)=\ell(y,\targetModel(x))$. In the original algorithm introduced by \cite{KurakinGB16,madry2018towards}, the step size $\eta^{(k)}$ is fixed, while Auto-PGD uses an adaptive step size which  improves the performance and makes the algorithm model-agnostic.

% Definition~\ref{def:adv_strat} can be extended if one considers an adversarial strategy that encompasses the whole class of classifier as detailed in  the following definition. %Definition~\ref{def:adv_strat_global}.

% \begin{definition}{(Untargeted Adversarial Strategy over $\mathcal{G}$)} An endomorphism $\mathcal{V}_\mathcal{G}$ defined on the input space $\mathcal{X}$ is an untargeted adversarial strategy over a class of classifiers $\mathcal{G}$, if for any $g \in \mathcal{G}$ and any $x \in \mathcal{X}$ there exists $v \in \mathcal{X}$, $ v\defeq \mathcal{V}_\mathcal{G}(x)$ --possibly not unique--, such that
%   \begin{equation}
%   g(x + v) \not = g(x).
% \end{equation}
% \label{def:adv_strat_global}
% \end{definition}

% Let $v^\star$ denote the orthogonal projection of an input $x$ into the manifold $\{x \in \mathcal{X}, g(x)=0\}$ \cite{moosavi2016deepfool}, $v^\star$ naturally corresponds to

% \begin{equation}
%   v^\star = \argmin_{v \in \mathcal{X} \text{ s.t. } g(v)=0} \| \ v - x \|_2^2
% \end{equation}

% Gradient based methods \cite{goodfellow2014explaining, biggio2018wild, ozbulak2020perturbation} may rely on a stochastic approximation of the gradients \cite{athalye2018obfuscated} to build adversarial examples from training data. Thus the resulting   

\section{Membership Inference Attacks from Adversarial Examples}
\label{sec:AdvDist}

% \label{sec:MembershipInferenceAttacksfromAdversarialExamples}
\label{sec:MembershipInferenceAttacksfromAdversarialExamples}
% Hereafter, we introduce novel strategies for MIA and explain their connection to adversarial attacks.

In this section, the core elements of this paper are discussed as we show how the adversarial distance bridges the gap between MIA and adversarial examples. We introduce our attack and  describe the resulting algorithm in detail.

% \subsection{Adversarial Distance}

% We shall consider a membership inference strategy $\phi_g : \mathcal{X} \mapsto \{0,1\} $ defined as follows, for any sample $x$ from the input space $\mathcal{X}$ and any classification model $g$:
% \begin{align*}
%   \phi_g(x) = \un\{\|v\| \geq \tau \}.
% \end{align*}
% where $v \in \mathcal{X}$ is a vector output following Definition~\ref{def:adv_strat} by $\mathcal{V}$ designed to fool the classifier $g$ \textit{i.e.} it verifies that $g(x + v) \not = g(x)$. This attack strategy guarantees that the 
% Auto-PGD maximizes the loss $\ell(y,\targetModel(x))$ with respect to the input sample $x$; since the same loss was minimized during training over the training set, maximizing the loss requires stronger perturbations for this samples.

During training, the target model minimizes the loss over samples from the training set. The objective of Projected Gradient Descent \cite{madry2018towards,KurakinGB16} and other algorithms derived from it (e.g., Auto-PGD \cite{AutoAttack}) is to maximize the very same loss. Hence, we expect this process to require larger perturbations for members of the training set, compared to samples that were not observed during training. We exploit this feature to perform  MIAs against machine learning models.

Our membership inference strategy measures the distance between an adversarial example and its original counterpart, i.e., the size of the perturbation, and uses this as a criteria to distinguish members of the training set,
\begin{align*}
    \varphi(\testSample,\targetModel) = \Vert \AdvStrat_{p,\epsilon}(\testSample,\targetModel) \Vert_p \;,
\end{align*}
where $\Vert \cdot \Vert_p$ measures the length of the perturbation. In our experiments, we use either $l_1$, $l_2$ or $l_\infty$ norm %to measure the distance between samples
(i.e., $p\in\{1,2,\infty\}$) and the same norm is used to constraint the size of the perturbation produced by the adversarial strategy, guaranteeing that $\varphi(\testSample,\targetModel) \le \epsilon$ (see \Cref{alg:algo1}).

\begin{figure}
    \centering
    \includegraphics[width=0.45\textwidth]{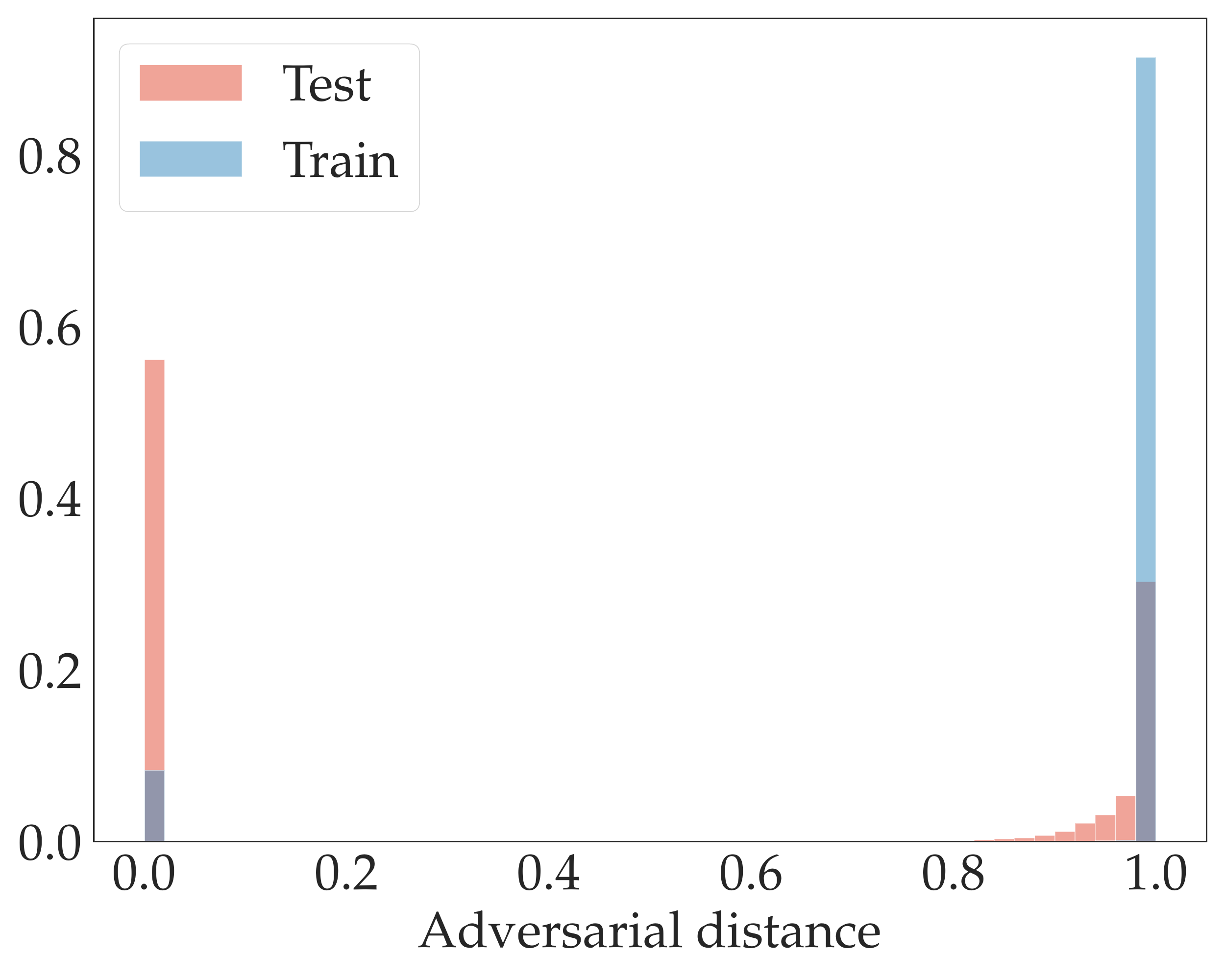}
    \caption{Histogram of adversarial distances over $50k$ samples from the training set (blue) superposed to the same histogram over $10k$ samples from the test set. The adversarial examples are computed for AlexNet, trained on CIFAR-100, based on the $\|\cdot\|_\infty$ norm and $\epsilon = 1$.}
    \label{fig:histogram}
\end{figure}

\begin{algorithm}[!h]
\caption{ }
\algsetup{linenodelimiter=.}
\begin{algorithmic}[1]
  \REQUIRE Target sample $\testSample$, target model $\targetModel$, adversarial strategy $\AdvStrat_{p,\epsilon}$, $p\in\{1,2,\infty\}$, $\epsilon>0$ and, $\tau\in\rset_+$.%
\STATE $v\leftarrow\AdvStrat_{p,\epsilon}(\testSample,\targetModel)$ %\COMMENT{Compute the adversarial perturbation provided by the adversarial strategy $\mathcal{V}$}.
\\
\textcolor{blue}{$\backslash\backslash$\texttt{Adversarial perturbation $v$.}}
%\STATE $d(x_\mathrm{test},x_\mathrm{test}+v)\leftarrow\Vert v\Vert_p$\\
%\RETURN $\un\{d(x_\mathrm{test},x_\mathrm{test}+v) \geq \tau\}$
\RETURN $\un\{\Vert v\Vert_p \geq \tau\}$
\\
\textcolor{blue}{$\backslash\backslash$\texttt{Is the distance between the adv. ex. and the original input $x_{\text{test}}$ greater than $\tau$?}} 
\end{algorithmic}
\label{alg:algo1}
\end{algorithm}
Since we are not interested in producing subtle perturbations that preserve the perspective of a human, we let the adversarial attacker generate arbitrarily large perturbations (constrained only by the dynamic range of the image). However, as shown in \cref{fig:histogram} and as demonstrated in the experimental section, there is a significant shift in the distribution of the size of perturbations, depending on whether (or not) the samples are part of the training set. 

\section{Review of MIAs Relying on Training Data}
Most of the MIA strategies proposed in the literature combine sets of features of the target sample and target model. Combining these features into a single score that can be used for a binary decision test is a challenging task. A common strategy is to train a machine learning model that learns to combine these features and predict whether the target sample belongs to the training set or not. Naturally, the attack model requires a set of samples that are labeled as either part of the training set of the target model or outside of the training set of the target model. In this section we present a short review of attack models previously proposed in the literature:

\noindent\textbf{Grad $x$ and Grad $w$ Attack Models}:
In \cite{Rezaei}, they propose an attack model that uses an array of statistics from the gradient of the loss of target model. The statistics considered are the $\ell_1$ norm, $\ell_2$ norm, maximum value, mean, skewness, kurtosis and absolute minimum of the gradient. These statistics are combined by a logistic regression model trained on labeled data. We implement this attack model and reproduce the results in our setting. When the gradient is taken with respect to the model parameters, we refer to the attack model as `Grad $w$'. On the other hand, when the gradient is taken with respect to the input sample, we refer to the attack model as `Grad $x$'.

\noindent\textbf{Intermediate Outputs}:
The authors of \cite{Rezaei} also consider an attack model that uses the intermediate outputs of the target model. For the models considered in this work, the attacker uses the outputs of the last two layers of the target model. This attack model is also implemented as described in the original paper and evaluated in our setting. This model is later abbreviated as Int. Outs.

\noindent\textbf{White-Box \cite{NasrShokri}}:
The attack model proposed by \cite{NasrShokri} utilizes the gradients of the loss function with respect to model parameters at the target sample, the value of the loss at the target sample, intermediate outputs of the target model and the one hot encoded labels of the target sample. To our knowledge this was the first work to propose using the gradient of the loss w.r.t. model parameters as a criteria to infer membership. We implement this attack strategy and reproduce the results presented in their paper. This attack strategy is referred to as WB \cite{NasrShokri}.

\noindent\textbf{Ensemble Attacker} We propose an ensemble attacker that takes as input the Softmax response of the target model, the value of its loss function, the norm of the gradient of the loss with respect to the model parameters, the norm of the gradient of the loss with respect to the input sample, the modified entropy, and the adversarial distance. This model outperforms the state-of-the-art against AlexNet, and achieves similar performance against ResNext. The details of the model are presented in the supplementary material \cref{sec:ensemble}. The results for this model are also detailed in the supplementary material \cref{sec:additionalresults}.

% \hamid{I suggest to place the description of the network in the Appendix.} 

% Similar to Softmax Response, this attack exploits the confidence of the model, but this time on adversarially perturbed samples. This attack is implemented for different perturbation budgets. The pertubations are generated in an ``untargeted" way, which means they attempt to change the prediction of the model to any other class.

% The targeted version of the previous attack generates adversarial examples for each class (other than the original one) and then these adversarial examples are input to the model, and the model's response for each of these is input to an attacker model that performs a binary prediction about the membership of the original sample.
% Exploiting verified worst-case predictions.

% They use the $\ell_{\infty}$ perturbation constraint throughout their experiements. They perform experiments on different datasets (Yale Face, FashionMNIST, CIFAR10) and use either ResNet or a CNN trained by them as target models.

%\section{Theoretical Analysis}
%\label{sec:theoreticalAnalysis}
%\input{theoreticalAnalysis}

%\section{A possible defense against designed attacks}
%\label{sec:defense}
%\input{defense}

\section{Numerical Experiments}
\label{sec:experiments}

% \begin{table*}[t]
%     \centering
%     \begin{tabular}{l|c|c|c|c|c}
%     Target & \multicolumn{2}{c|}{Black Box} & \multicolumn{2}{c|}{White Box} & Side Information\\
%     \cline{2-6}
%     Model & Softmax & Loss & Gradient Norm & Adv. Distance & {\MLattacker} \\
%     \hline
%     AlexNet & $67.9\pm0.2$ & $76.9\pm0.1$ & $76.4\pm0.2$ & $84.4\pm0.2$ & $\mathbf{88.8\pm0.3}$ \\
%     ResNet & $55.5\pm0.1$ & $58.8\pm0.1$ &$60.0\pm0.1$ & $84.7\pm0.1$ & $\mathbf{88.8\pm0.4}$ \\
%     ResNext & $72.3\pm0.1$ & $72.5\pm0.1$ &$73.0\pm0.1$ & $89.2\pm0.0$ & $\mathbf{92.3\pm0.3}$ \\
%     DenseNet & $70.5\pm0.1$ & $70.8\pm0.1$ &$71.3\pm0.1$ & $83.1\pm0.1$ & $\mathbf{87.5\pm0.3}$ \\
%     \end{tabular}
%     \caption{AUROC($\%$). Comparison of different MIA Strategies.}
%     \label{tab:AUROC1}
% \end{table*}

Hereafter, we first present the experimental setting for MIAs and then provide numerical results on real world data. The code necessary to reproduce these experiments is available in our repository\footnote{\url{https://github.com/ganeshdg95/Leveraging-Adversarial-Examples-to-Quantify-Membership-Information-Leakage}}. Further intuition and results are presented in the supplementary material \cref{sec:appendix_details}.

\begin{table*}[t]
    \centering
    \resizebox{\textwidth}{!}{
    \begin{tabular}{l|c|c|c|c|c|c|c|c}
    MIA Strategy & \multicolumn{2}{c|}{AlexNet} & \multicolumn{2}{c|}{ResNet} & \multicolumn{2}{c|}{ResNext} & \multicolumn{2}{c}{DenseNet} \\
    \cline{2-9}
    \textcolor{red}{} & AUROC & Accuracy & AUROC & Accuracy & AUROC & Accuracy & AUROC & Accuracy  \\
    \hline
    Softmax & $68.00\pm0.16$ & $65.34\pm0.14$ & $55.45\pm0.15$ & $57.40\pm0.13$ & $72.37\pm0.07$ & $74.84\pm0.11$ & $70.52\pm0.09$ & $72.11\pm0.10$ \\
    Mentr. \cite{Song2021} & $77.11\pm0.10$ & $74.16\pm0.11$ & $59.10\pm0.13$ & $61.39\pm0.11$ & $76.87\pm0.08$ & $75.28\pm0.11$ & $74.21\pm0.10$ & $72.69\pm0.10$ \\
    Loss & $76.69\pm0.10$ & $74.14\pm0.13$ & $58.66\pm0.13$ & $61.29\pm0.11$ & $72.57\pm0.07$ & $75.17\pm0.11$ & $70.85\pm0.09$ & $72.61\pm0.10$ \\
    Grad Norm & $76.58\pm0.10$ & $74.19\pm0.12$ & $59.93\pm0.13$ & $62.56\pm0.09$ & $73.06\pm0.07$ & $75.74\pm0.11$ & $71.30\pm0.09$ & $73.81\pm0.09$ \\
    Adv. Dist. (ours) & $\pmb{84.35\pm0.13}$ & $\pmb{85.12\pm0.18}$ & $\pmb{84.53\pm0.16}$ & $\pmb{85.45\pm0.11}$ & $\pmb{89.24\pm0.03}$ & $\pmb{89.10\pm0.05}$ & $\pmb{82.76\pm0.03}$ & $\pmb{82.63\pm0.05}$ \\
    \hline
    \hline
    Grad $w$* \cite{Rezaei} & $78.76\pm0.30$ & \pmb{$74.32\pm0.28$} & $61.98\pm0.38$ & $62.72\pm0.27$ & $77.80\pm0.30$ & $73.47\pm0.57$ & $73.12\pm1.42$ & $72.59\pm0.55$ \\
    Grad $x$* \cite{Rezaei} & $77.20\pm0.26$ & $73.43\pm0.26$ & $68.48\pm0.27$ & $63.58\pm0.22$ & $77.54\pm0.61$ & $73.47\pm0.57$ & $75.81\pm0.43$ & $71.81\pm0.40$ \\
    Int. Outs* \cite{Rezaei} & $57.92\pm0.50$ & $56.36\pm0.41$ & $\pmb{96.59\pm0.29}$ & $\pmb{91.57\pm0.43}$ & $\pmb{93.62\pm0.39}$ & \pmb{$86.38\pm0.37$} & $\pmb{99.17\pm0.10}$ & $\pmb{97.68\pm0.14}$ \\
    WB* \cite{NasrShokri} & \pmb{$80.33\pm1.21$} & $74.03\pm0.71$ & $87.51\pm0.41$ & $79.73\pm0.30$ & $84.52\pm1.95$ & $76.46\pm1.82$ & $79.38\pm1.16$ & $71.92\pm0.97$ \\
    % {\MLattacker}* (ours) & $\pmb{90.84\pm0.13}$ & $\pmb{85.48\pm0.67}$ & $89.31\pm1.04$ & $85.07\pm0.47$ & $92.30\pm0.19$ & $\pmb{92.17\pm0.15}$ & $87.86\pm0.22$ & $87.46\pm0.20$ \\
    \end{tabular}
    }
    \caption{Comparison of different MIA Techniques. The Accuracy($\%$) and AUROC score ($\%$) on a balanced evaluation set are reported. $10k$ are uniformly selected from the training set (members) and the whole $10k$ samples from the testing set are selected (non-members). All the data selected is used for evaluation. Techniques with a (*) require training. In this case, only $60\%$ of the data is used for evaluation and rest is used for training. %\textcolor{red}{\# Add bold letters on the best performances.\#}
    }
    
    \label{tab:AUROC}
\end{table*}
% without data one can still perform the MIA. 

\subsection{CIFAR10 and CIFAR100 Datasets}
\label{sec:datasets}
%\noindent\textbf{CIFAR100}. 
These datasets are a standard benchmark for image recognition tasks \cite{Krizhevsky09learningmultiple}. They contain $60000$ $32\times32$ pixels, color (RGB) images split amongst $10$, $100$ distinct classes, respectively. In standard libraries, like PyTorch \cite{NEURIPS2019_9015}, these datasets are usually divided into a training set containing $50k$ images and a test set containing the remaining $10k$ images. The standard training set provided by PyTorch is used to train the target models we consider, the rest is used as outside-the-training-set data.
\subsection{Target Models}

\noindent\textbf{State-of-the-art models for image recognition}. We consider popular models for image recognition, pre-trained and publicly available\footnote{Model implementations, pre-trained weights and code to train the models available at \url{https://github.com/bearpaw/pytorch-classification}}. Namely, the models considered are AlexNet \cite{krizhevsky2012imagenet}, ResNet \cite{he2016deep}, ResNext \cite{xie2017aggregated} and DenseNet \cite{huang2017densely}, trained for image classification on CIFAR-100. These are the same pretrained models considered in \cite{NasrShokri,Rezaei}. 
\subsection{Comparison of MIA strategies}
%\label{sec:experiments}
\begin{figure*}[!h]
        \centering
        \begin{subfigure}[b]{0.24\textwidth}
            \centering
            \includegraphics[width=\textwidth]{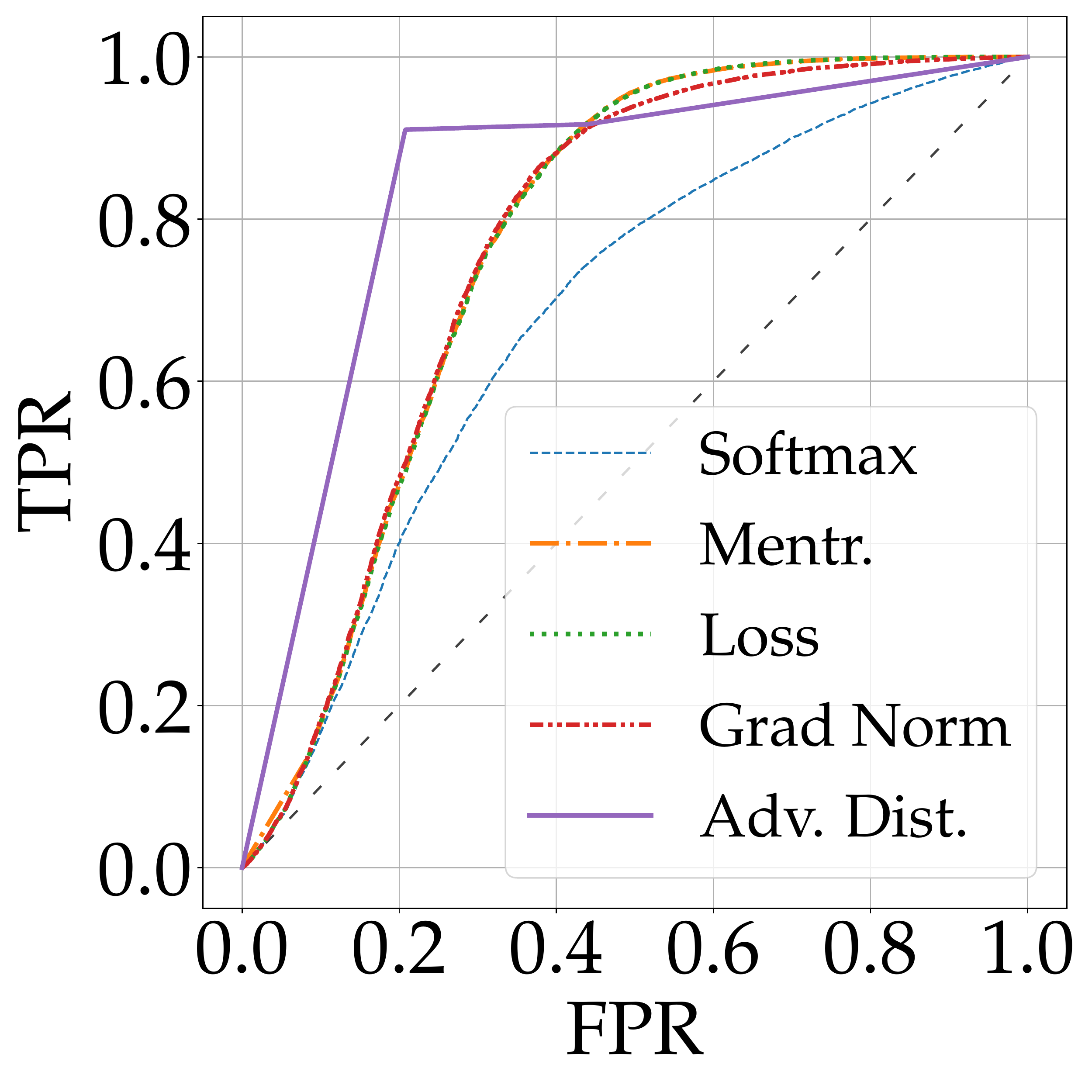}
            \caption[Network2]%
            {{\small AlexNet}}    
            \label{fig:AlexNet}
        \end{subfigure}
        \begin{subfigure}[b]{0.24\textwidth}  
    \includegraphics[width=\textwidth]{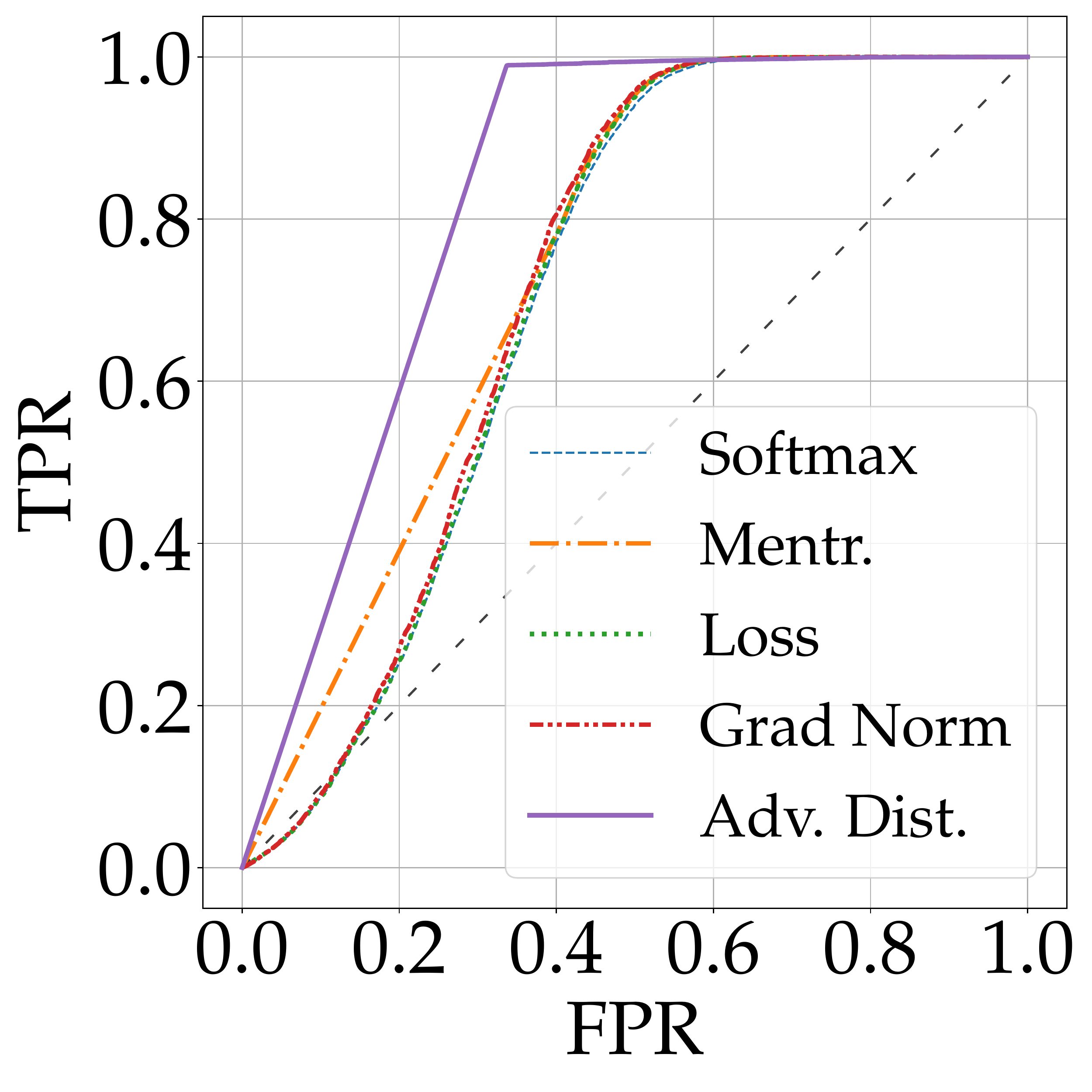}
            \caption[]%
            {{\small DenseNet}}    
            \label{fig:DenseNet}
        \end{subfigure}
        %\vskip\baselineskip
        \begin{subfigure}[b]{0.24\textwidth}   
            \centering  \includegraphics[width=\textwidth]{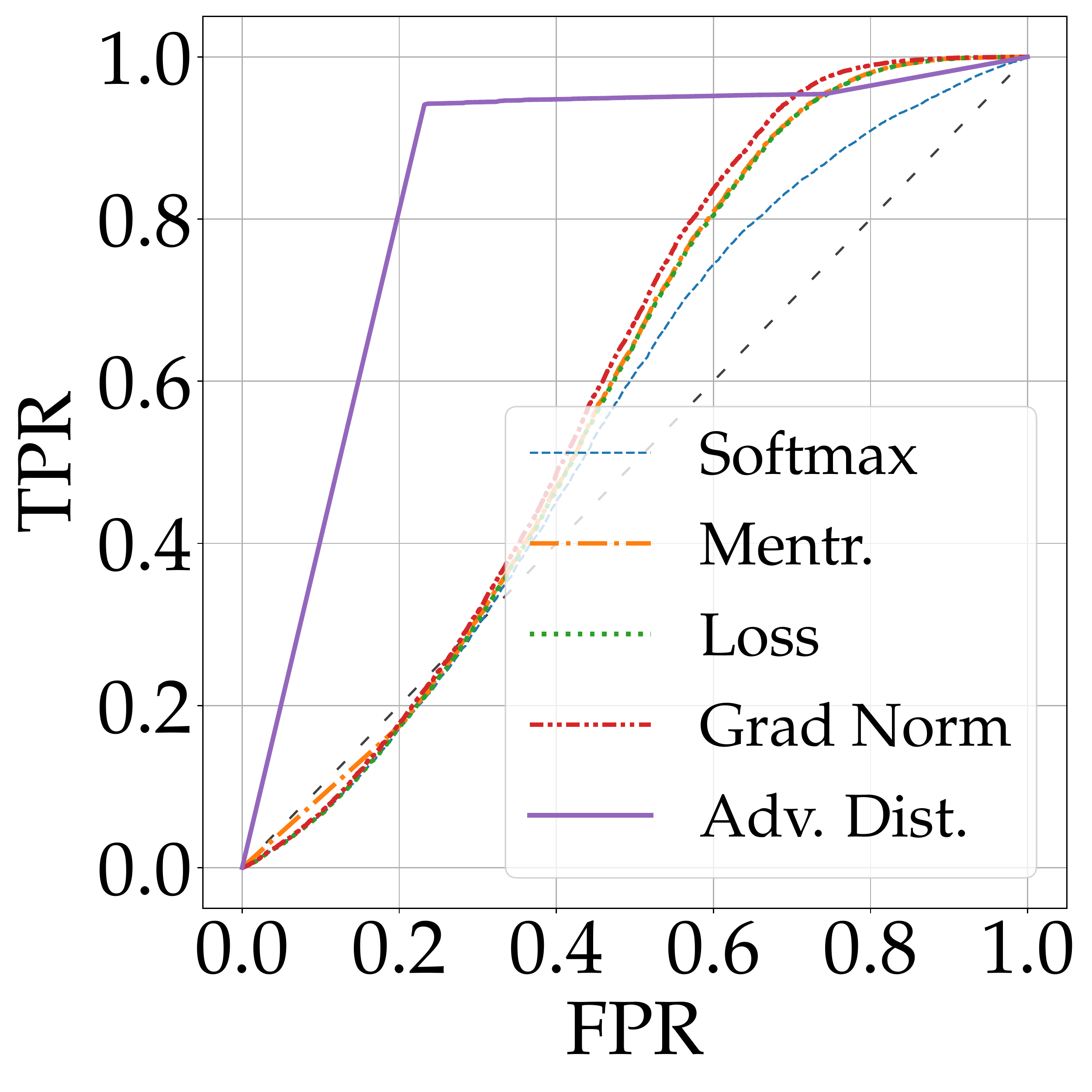}
            \caption[]%
            {{\small ResNet}}    
            \label{fig:ResNet}
        \end{subfigure}
        %\hfill
        \begin{subfigure}[b]{0.24\textwidth}   
            \centering 
 \includegraphics[width=\textwidth]{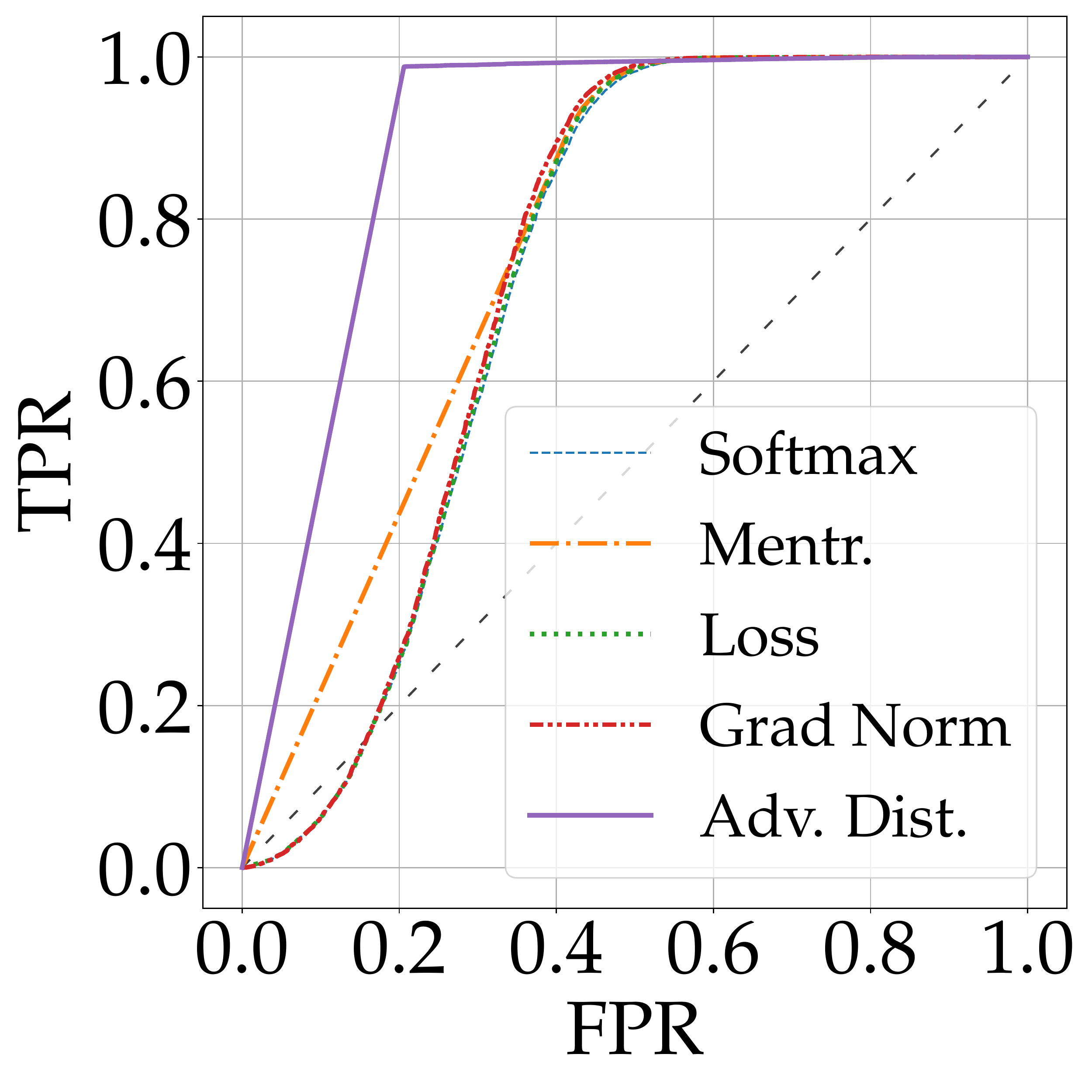}
            \caption[]%
            {{\small ResNext}}    
            \label{fig:Resnext}
        \end{subfigure}
        \caption[ The average and standard deviation of critical parameters ]
        {\small ROC curves of different MIA strategies against AlexNet (\ref{fig:AlexNet}), 
        DenseNet (\ref{fig:DenseNet}),
        ResNet (\ref{fig:ResNet}),
        ResNext (\ref{fig:Resnext}). These are computed on a balanced evaluation set and averaged over $20$ iterations, as described for \cref{tab:AUROC}.}
        \label{fig:rocs}
\end{figure*}

To evaluate a membership inference strategy, two groups of samples are needed: samples from the training set and samples outside the training set of the target model. The pre-trained models considered in this work are trained on $50k$ samples from the CIFAR-100 dataset. The remaining $10k$ samples constitute the test set. 

We perform membership inference attacks using nine different strategies. The Softmax Response, Modified Entropy and Loss strategies are black-box strategies, since the attacker only requires access to the target sample, its label and the output of the model (either the logits or Softmax response of the model). On the other hand, the Gradient Norm and Adversarial Distance strategies are white-box, as the attacker requires access to the model parameters in order to compute gradients of the loss function. % with respect to these
% (Gradient norm, Adversarial Distance). % or with respect to the input sample ().
In addition to white-box access to the target model, for some of the strategies we consider, the attacker requires its own training set of samples, including a subset of the training set used by the target model. This is the case for the Grad $w$, Grad $x$, Intermediate Outputs (Int. Outs) and the White-Box (WB) attacker.

% We compare our results to those of \cite{NasrShokri}. In Table~\ref{tab:SOTAComp} we report the accuracy achieved by all the strategies proposed in the paper against different models. The threshold is chosen along the ROC curve to maximize accuracy. The target models considered are the same pre-trained models, taken from the same publicly available repository as our competitors. We compare our strategy that requires side information to those of  \cite{NasrShokri}. The results shown on the ``BB'' and ``WB'' columns of \cref{tab:SOTAComp} are cited from the ``black-box" and ``white-box'' columns from table~VIII in their paper, respectively. We highlight the best performing strategy against each model with boldface. Remark that the attackers from \cite{NasrShokri} require a training set of their own, while most of the attack strategies we consider here do not require any extra information. With this in mind, it is remarkable that the Adversarial Distance attacker significantly and consistently outperforms their strategies across all target models, and in some cases also outperforms our {\MLattacker}. Additionally, note that the Softmax Response outperforms their black-box attacker for the case of DenseNet, but falls short on the other cases. We observe that additional knowledge of samples in the training set might improve the performance of the attacker, but it is not necessary in order to observe information leakage from the model.

\begin{table*}[t]
    \centering
    \resizebox{\textwidth}{!}{
    \begin{tabular}{l|c|c|c|c|c|c|c|c}
     MIA Strategy & \multicolumn{2}{c|}{AlexNet} & \multicolumn{2}{c|}{ResNet} & \multicolumn{2}{c|}{ResNext} & \multicolumn{2}{c}{DenseNet} \\
    \cline{2-9}
     & Accuracy & FPR & Accuracy & FPR & Accuracy & FPR & Accuracy & FPR \\
    \hline
    Softmax & $65.17\pm0.48$ & $43.01\pm1.00$ & $57.32\pm0.43$ & $63.64\pm0.81$ & $74.97\pm0.52$ & $45.82\pm0.99$ & $71.98\pm0.51$ & $51.25\pm1.00$ \\
    Mentr. \cite{Song2021} & $74.04\pm0.46$ & $40.96\pm1.23$ & $61.38\pm0.30$ & $67.43\pm0.84$ & $75.44\pm0.56$ & $44.59\pm1.12$ & $72.59\pm0.53$ & $50.47\pm1.03$ \\
    Loss & $74.00\pm0.47$ & $40.88\pm1.20$ & $61.28\pm0.34$ & $67.95\pm0.63$ & $75.31\pm0.54$ & $45.15\pm1.04$ & $72.49\pm0.51$ & $50.67\pm0.84$ \\
    Grad Norm & $74.10\pm0.52$ & $39.19\pm1.14$ & $62.55\pm0.31$ & $67.81\pm0.83$ & $75.81\pm0.58$ & $43.78\pm0.97$ & $73.06\pm0.50$ & $49.56\pm0.95$ \\
    Adv. Dist. (ours) & $\pmb{85.21\pm0.45}$ & $\pmb{20.60\pm0.87}$ & $\pmb{85.56\pm0.36}$ & $\pmb{22.98\pm0.77}$ & $\pmb{89.19\pm0.35}$ & $\pmb{20.39\pm0.70}$ & $\pmb{82.78\pm0.49}$ & $\pmb{33.37\pm0.97}$ \\
    \hline
    \hline
    Grad $w$* \cite{Rezaei} & $68.35\pm0.50$ & $60.74\pm1.04$ & $59.98\pm0.34$ & $78.86\pm0.67$ & $66.59\pm0.42$ & $66.74\pm0.86$ & $64.54\pm0.37$ & $70.87\pm0.72$ \\
    Grad $x$* \cite{Rezaei} & $62.84\pm0.45$ & $71.18\pm1.00$ & $59.76\pm0.43$ & $79.51\pm0.85$ & $65.37\pm0.48$ & $69.22\pm0.96$ & $63.48\pm0.34$ & $73.01\pm0.69$ \\
    Int. Outs* \cite{Rezaei} & $53.90\pm0.45$ & $77.05\pm2.43$ & $\pmb{87.97\pm1.34}$ & $\pmb{19.92\pm3.03}$ & $\pmb{85.67\pm0.91}$ & $\pmb{24.56\pm2.32}$ & $\pmb{96.45\pm0.88}$ & $\pmb{5.63\pm1.85}$ \\
    WB* \cite{NasrShokri} & $\pmb{70.76\pm0.80}$ & $\pmb{56.54\pm1.91}$ & $65.75\pm3.48$ & $65.90\pm7.70$ & $69.33\pm2.12$ & $59.68\pm4.81$ & $65.35\pm2.07$ & $68.25\pm4.64$ \\
    % {\MLattacker}* (ours) & $\pmb{81.27\pm1.97}$ & $\pmb{35.31\pm4.90}$ & $80.47\pm2.24$ & $37.13\pm4.80$ & $\pmb{84.23\pm5.98}$ & $30.71\pm1.22$ & $78.34\pm5.48$ & $42.73\pm11.20$
    \end{tabular}
    }
    \caption{Comparison of different MIA Techniques. The balanced accuracy ($\%$) and FPR ($\%$) on an imbalanced evaluation set are reported. The whole training set ($50k$ samples) and testing set ($10k$ samples) are used; thus ratio between members and non-members is 5:1. $80\%$ of the total data is used to find the threshold that maximizes accuracy and the other $20\%$ is used to evaluate this threshold. Techniques with a (*) require training. In this case, instead of using $80\%$ of the total data to determine the threshold, this $80\%$ is used for training and the threshold is set at $0.5$.} 
    %\textcolor{red}{\# SAME COMMENT AS IN PREVIOUS TABLES. \#}
    \label{tab:RezaeiAnalysis}
\end{table*}

% \begin{table*}[!h]
%     \centering
%     \begin{tabular}{l|c|c|c|c|c|c|c}
%     Target & \multicolumn{2}{c|}{Black Box} & \multicolumn{2}{c|}{White Box} & \multicolumn{3}{c}{Side Information}\\
%     \cline{2-8}
%     Model & Softmax & Loss & Gradient Norm & Adv. Distance & BB \cite{NasrShokri} & WB \cite{NasrShokri}& {\MLattacker} \\
%     \hline
%     AlexNet & $65.1\pm0.2$ & $74.2\pm0.2$ & $74.2\pm0.2$ & $\mathbf{85.1\pm0.2}$& $74.2$ & $75.1$ & $83.2\pm0.7$ \\
%     ResNet & $57.5\pm0.1$ & $61.4\pm0.1$ &$62.6\pm0.1$ & $\mathbf{85.6\pm0.1}$ & $62.2$ & $64.3$ & $84.1\pm0.3$ \\
%     ResNext & $74.8\pm0.1$ & $75.2\pm0.1$ &$75.7\pm0.1$ & $89.1\pm0.0$& - & - & $\mathbf{92.1\pm0.5}$ \\
%     DenseNet & $72.1\pm0.1$ & $72.6\pm0.1$ &$73.1\pm0.1$ & $83.0\pm0.1$ & $67.7$ & $74.3$ & $\mathbf{87.6\pm0.2}$ \\
%     \end{tabular}
%     \caption{Best Accuracy($\%$). Comparison to Nasr et al.}
%     \label{tab:SOTAComp}
% \end{table*}

In our analysis we consider a balanced evaluation set and report the AUROC score and the maximum accuracy achieved by each strategy. In this setting, a subset of $10k$ samples from the training set is selected  uniformly as \textit{in-training} data and the entire test set ($10k$ samples) is selected as \textit{out-of-training} data. Since the choice of the subset of the training set influences our results, the experiments are repeated $20$ times, choosing a different subset each time. All the quantities reported are averaged over these $20$ runs of the experiment and the error reported is the empirical standard deviation. The results of this analysis are reported in \cref{tab:AUROC}. For each target model, the best performing attack strategies are highlighted in boldface. Note that the upper part of the table corresponds to strategies that do not require to train an attack model, nor require any additional samples, while the bottom part corresponds to strategies that require training an attack model. The best performing strategy when no additional samples are available is the adversarial distance strategy. Note that this strategy performs consistently across all target models and even surpasses the more resource hungry strategies for the case of AlexNet and ResNext. When additional samples are available, the Intermediate Outputs strategy is the most effective against the ResNet and DenseNet models. It is worth to mention that it might be infeasible for the attacker to obtain enough samples from the training set of the target model to launch this attack, in which case Adversarial Distance might be a better alternative. \Cref{fig:rocs} shows the ROC curves for the different strategies mentioned in the upper portion of \cref{tab:AUROC}. Note the jump on the ROC curve of the adversarial distance strategy. Due to a large group of samples having a score of $0$, increasing the threshold slightly above results in a sudden increase in the true positive rate (TPR). Indeed, this strategy takes advantage of the fact that most misclassified samples lie outside of the training set. 

Additionally, to demonstrate that our results are not biased by our analysis we perform the same analysis as in \cite{Rezaei}. In this setting the whole training set ($50k$ samples) and the whole test set ($10k$ samples) are used, resulting in an unbalanced evaluation set. To alleviate this issue, the metrics considered are robust to the training:test sample ratio. Namely, the False Positive Ratio (FPR) and balanced accuracy are considered. Table~\ref{tab:RezaeiAnalysis} presents the results of this analysis. Note that these results are consistent with \cref{tab:AUROC}; in particular, Adversarial Distance remains the best performing strategy across all models when no additional samples are available, also outperforming the state-of-the-art against AlexNet and ResNext. On the other hand, Intermediate Outputs is the best option against ResNet and DenseNet out of the models that require additional information and training.

To demonstrate that the effectiveness of our approach is not limited to one dataset, we repeat our experiments on CIFAR10. The target model considered is AlexNet and both the analyses from \cref{tab:AUROC} and from \cref{tab:RezaeiAnalysis} are considered. The results, shown in \cref{tab:cifar10}, are consistent with the results for CIFAR100, in the sense that our strategy remains the most effective. However, the performance of MIAs in general is worsen in comparison to CIFAR100. This is tied to the fact that the classification problem at hand has less classes, a trend that was previously observed in the literature \cite{truex2019demystifying}.
\begin{table}[t]
    \centering
    \resizebox{0.5\textwidth}{!}{
    \begin{tabular}{l|c|c|c|c}
     & \multicolumn{2}{c|}{Analysis 1} & \multicolumn{2}{c}{Analysis 2} \\
    \cline{2-5}
    \textcolor{red}{} MIA Strategy & AUROC & Accuracy & FPR & Accuracy  \\
    % \cline{2-5}
    %  & \multicolumn{4}{c}{AlexNet (CIFAR-10)} \\
    \hline
    Softmax & $54.24\pm0.21$ & $54.17\pm0.14$ & $60.94\pm1.01$ & $54.13\pm0.50$  \\
    Mentr. \cite{Song2021} & $57.05\pm0.22$ & $56.96\pm0.17$ & $69.58\pm0.96$ & $56.85\pm0.45$  \\
    Loss & $56.99\pm0.27$ & $56.90\pm0.17$ & $70.25\pm1.59$ & $56.77\pm0.45$ \\
    Grad Norm & $56.91\pm0.20$ & $56.95\pm0.17$ & $68.30\pm0.86$ & $56.91\pm0.46$ \\
    Adv. Dist. (ours) & $\pmb{77.78\pm0.24}$ & $\pmb{80.03\pm0.17}$ & $\pmb{29.36\pm0.69}$ & $\pmb{80.03\pm0.36}$ \\
    \hline
    \hline
    Grad $w$* \cite{Rezaei} &\pmb{$60.36\pm0.34$} & \pmb{$57.62\pm0.26$} & $97.61\pm0.29$ & $51.03\pm0.14$ \\
    Grad $x$* \cite{Rezaei} & $60.08\pm0.33$ & $57.53\pm0.34$ & $99.51\pm0.12$ & $50.18\pm0.06$ \\
    Int. Outs* \cite{Rezaei} & $58.65\pm0.58$ & $56.65\pm0.42$ & $\pmb{80.38\pm2.93}$ & $\pmb{54.34\pm0.72}$  \\
    WB* \cite{NasrShokri} & $55.57\pm4.77$ & $54.29\pm3.48$ & $92.17\pm1.43$ & $53.47\pm0.60$ \\
    \end{tabular}
    }
    \caption{Comparison of different MIA techniques against the AlexNet model trained on CIFAR10. The best accuracy ($\%$) and AUROC score ($\%$) on a balanced evaluation set are reported (marked as ``Analysis 1''). The best balanced accuracy ($\%$) and FPR ($\%$) at the same threshold on an unbalanced evaluation set are also computed (marked as ``Analysis 2''). Techniques with a (*) require training.
    }
    \label{tab:cifar10}
\end{table}
Finally, we test how robust the AUROC score is to a change in the proportion of training/testing samples. For this purpose we perform membership inference attacks using the Adversarial Distance strategy, and measure the AUROC score in three different settings, where we change the training:testing ratio. The results presented in \cref{tab:AdvDist} show that the AUROC score measures the effectiveness of the attack consistently, regardless of the ratio of members to non-members present in the evaluation set.

\begin{table}[h]
    \centering
    \resizebox{.45\textwidth}{!}{%
    \begin{tabular}{l|c|c|c}
    Target & \multicolumn{3}{c}{Training:Test ratio} \\
    \cline{2-4}
    Model & 5:1 & 1:1 & 1:5 \\
    \hline
    AlexNet & $84.36$ & $84.35\pm0.13$ & $84.39\pm0.41$  \\
    ResNet & $84.55$ & $84.53\pm0.16$ & $84.36\pm0.34$  \\
    ResNext & $89.24$ & $89.24\pm0.03$ & $89.24\pm0.08$  \\
    DenseNet & $82.76$ & $82.76\pm0.03$ & $82.75\pm0.08$  \\
    \end{tabular}%
    }
    \caption{Influence of evaluation set on performance for the Adversarial Distance strategy. The AUROC ($\%$) for different evaluation sets is reported. When the ratio is 5:1, the whole training is selected. When the ratio is 1:1, $10k$ samples from the training set are uniformly selected. When the ratio is 1:5, $2k$ samples from the training set are uniformly selected. In all cases the whole test set is also selected for evaluation.}
    \label{tab:AdvDist}
\end{table}

%\hamid{I commented out the Conclusion.}
\section{Summary and Concluding Remarks}
\label{sec:conclusion}
We have proposed a novel strategy for membership inference that pushes forward the state of the art. We have compared our  strategy to the best performing membership inference attacks in the literature using the same setting (standalone scenario with popular image classification models on CIFAR-100) and found that our strategy is the most consistent across all target models. Despite not using additional resources (no additional samples from the training set of the target model and no computational resources to train an attacker), our model outperforms the more resource demanding state-of-the-art methods against AlexNet and ResNext, and remains relevant in other cases.

We found that MIAs are highly effective against machine learning models and even without additional knowledge of samples from the training set of the target model, MIAs can reliably distinguish training samples from non-training ones (above $82\%$ accuracy regardless of the target model considered).

% We study how necessary it is to have additional samples from the training set of the target model in order to launch MIAs, 
% our strategy achieves comparable performance, without the need of  and without training an attack model. This makes our strategy more realistic and easy to transfer to any target model. Furthermore, our strategy is the most consistent across different target models, and even surpasses the SOTA against some of them.

\subsection{Limitations}

In this work we only consider target models trained on CIFAR10 and CIFAR100 as there is a lack of standard pre-trained models available for other data sets. Thence, comparing MIA strategies on such datasets is more difficult. In the future, we will extend our work to target models trained on other datasets.
% , and make the trained target models available for comparison.

On the other hand, always using the same pre-trained model for a particular architecture might result in a biased assessment of the privacy risks of that given architecture. Thus, it is equally important to evaluate the potential privacy risks on different shots of training, over the same dataset with the same architecture.

In this paper, we focused on the standalone scenario, \textit{i.e.} the target model is trained by a single entity which has access to the whole training set and the attacker does not observe the model during training.  This setup was chosen since it is the simplest setting. The ideas in this work can be extended to other scenarios (\textit{e.g.}, federated learning). 

\subsection{Discussion of (Potential) Negative Impact}

%\textcolor{red}{THIS SHOULD BE COMPETED: e.g., a bad guy can use this attack to reveal sensitive information from models currently deployed in nature ... }\\
%\hamid{
This article describes a novel attack strategy to retrieve information from the training data from any classification model. A direct potential negative impact of the work would be the improvement of attacks against machine learning models in production. Nevertheless, reliable and effective MIA strategies are needed to assess a model's privacy risks.
%}
% while not presenting possible mechanism of defense against Algorithm~\ref{alg:algo1}.

\section{Acknowledgement}

This research was supported by DATAIA ``Programme d'Investissement d'Avenir'' (ANR-17-CONV-0003) and by the ERC project Hypatia under the European Unions Horizon 2020 research and innovation program. Grant agreement No. 835294.

%%%%%%%%% REFERENCES
\clearpage
\newpage
\bibliographystyle{ieee_fullname}
\bibliography{biblio}

\newpage
\onecolumn
\appendix
\begin{center}
{\large {\bf\textsc{Supplementary material: Leveraging Adversarial Examples to Quantify \\ Membership Information Leakage
}}}
\end{center}

\Cref{sec:ensemble} describes our proposed ensemble membership inference attacker. In \cref{sec:appendix_details} we provide further experimental details and complementary results to those presented in the main paper, including results for several attack strategies that were initially considered, but under-performed. Finally, \cref{sec:additionalresults} provides additional results for the ensemble membership attacker.

\section{Ensemble Attacker}
\label{sec:ensemble}

This attacker requires not only white-box access to the model, as it needs to compute gradients with respect to input and to model parameters, but it also requires a training set of its own (similarly to  \cite{NasrShokri,shokri2017membership,Rezaei}). Essentially, what the attacker learns is how to map different observations to a membership label.

The attack model is a DNN with $5$ fully connected layers with output sizes $40,40,20,10$ and $1$, respectively. The input to the network is a vector of length $6$, containing the softmax response, modified entropy, loss value, gradient norm w.r.t.\ parameters, gradient norm w.r.t.\ input, and adversarial distance. These quantities are re-scaled to $[0,1]$, which significantly improves the performance of the model. The rescaling is done according to the maximum and minimum values from the training set. The model is trained with Adam optimizer \cite{Adam}  for up to $300$ epochs. The performance of the ensemble attacker is evaluated and compared to the performance of other strategies in \cref{tab:AUROC1}. Additionally, we vary the size of the attacker's training set and observe how this affects its performance. This results are presented in \cref{tab:MLattacker}.

\section{Further Experimental Details and Results}
\label{sec:appendix_details}

\subsection{Experimental Details}

Most of the experiments were run on a cluster with multiple nodes, each with NVIDIA Quadro RTX 6000 GPUs and an AMD EPYC 7302 16-Core processors.

When computing adversarial examples, we rescale the images so that their dynamic range lies within $[0,1]$. This is necessary in order for the adversarial attacks to compute distance and perform clipping properly. However, since the pre-trained models were trained on the natural images (previous to rescaling), we include an additional layer at the input of each target model that reverts the scaling, preserving the performance of the target model.

The accuracy presented in \cref{tab:AUROC} and \cref{tab:AUROC1} is computed by choosing a threshold along the ROC curve for each strategy. The threshold is chosen in order to maximize the accuracy. A similar process is done in the case of \cref{tab:RezaeiAnalysis}, where $80\%$ of the data is used to determine the threshold that maximizes the accuracy, and then the accuray is reported for the other $20\%$ of the data.

\subsection{Additional Results}

\begin{figure*}[!h]
        \centering
        \begin{subfigure}[b]{0.24\textwidth}
            \centering
            \includegraphics[width=\textwidth]{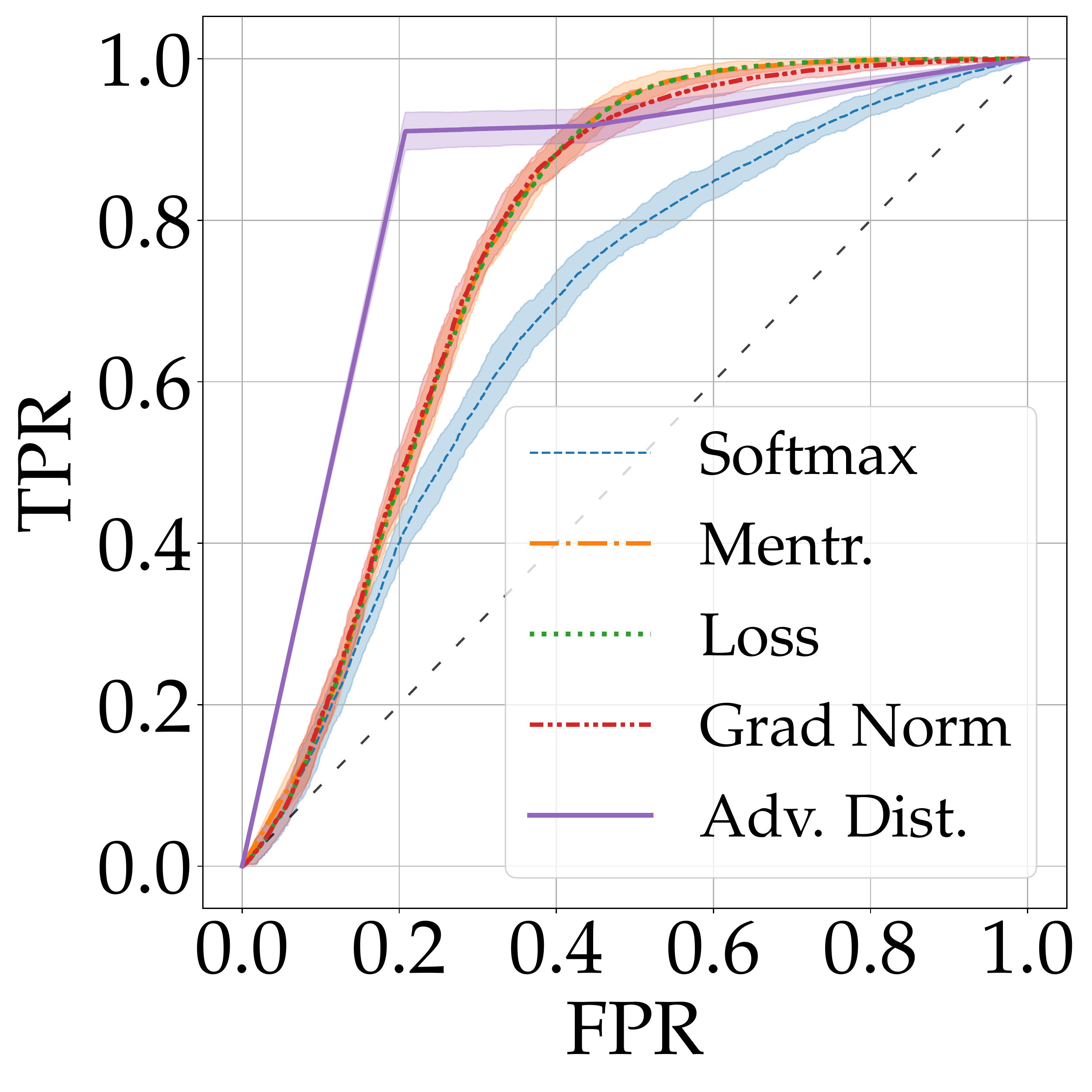}
            \caption[Network2]%
            {{\small AlexNet}}    
            \label{fig:AlexNet_}
        \end{subfigure}
        \begin{subfigure}[b]{0.24\textwidth}  
    \includegraphics[width=\textwidth]{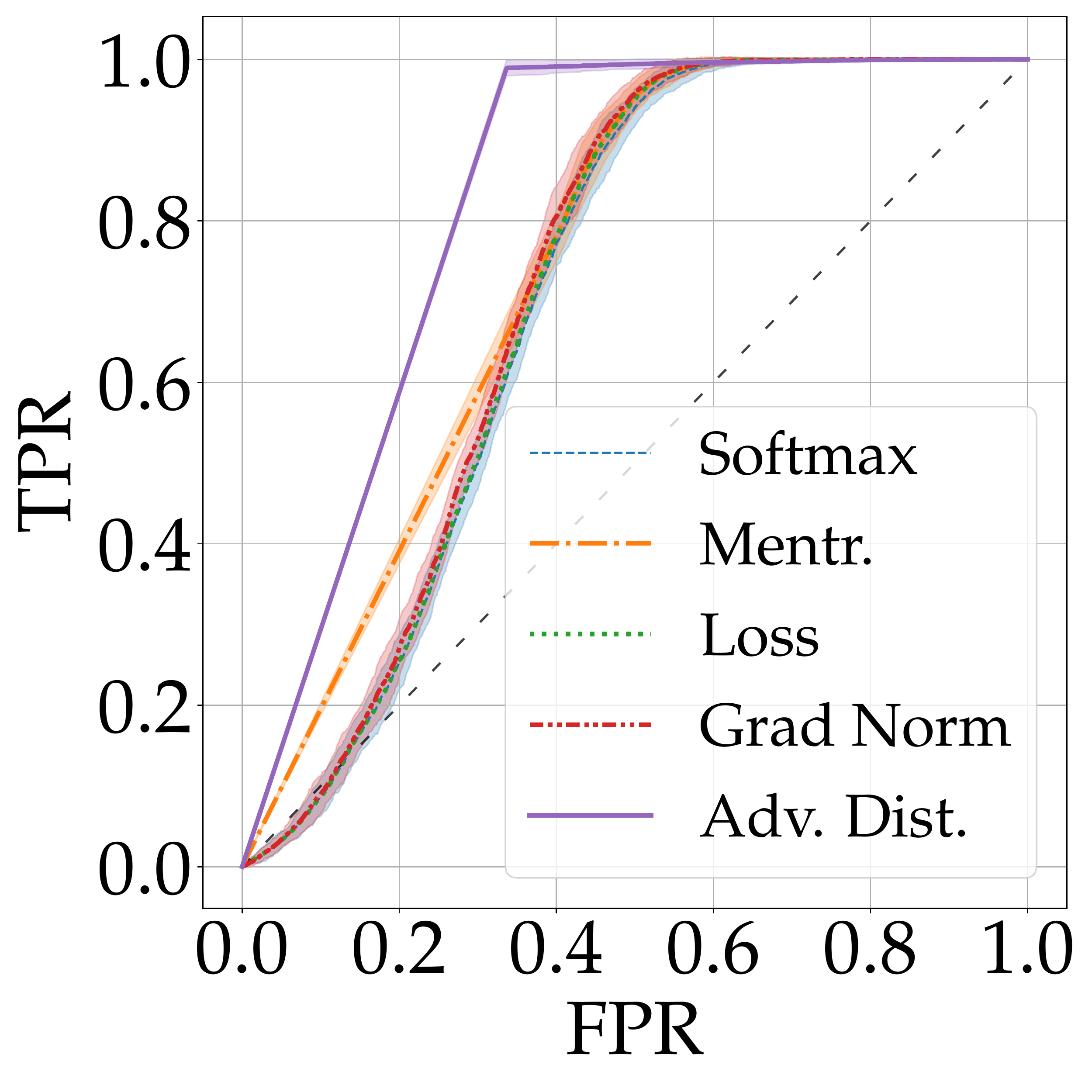}
            \caption[]%
            {{\small DenseNet}}    
            \label{fig:DenseNet_}
        \end{subfigure}
        %\vskip\baselineskip
        \begin{subfigure}[b]{0.24\textwidth}   
            \centering  \includegraphics[width=\textwidth]{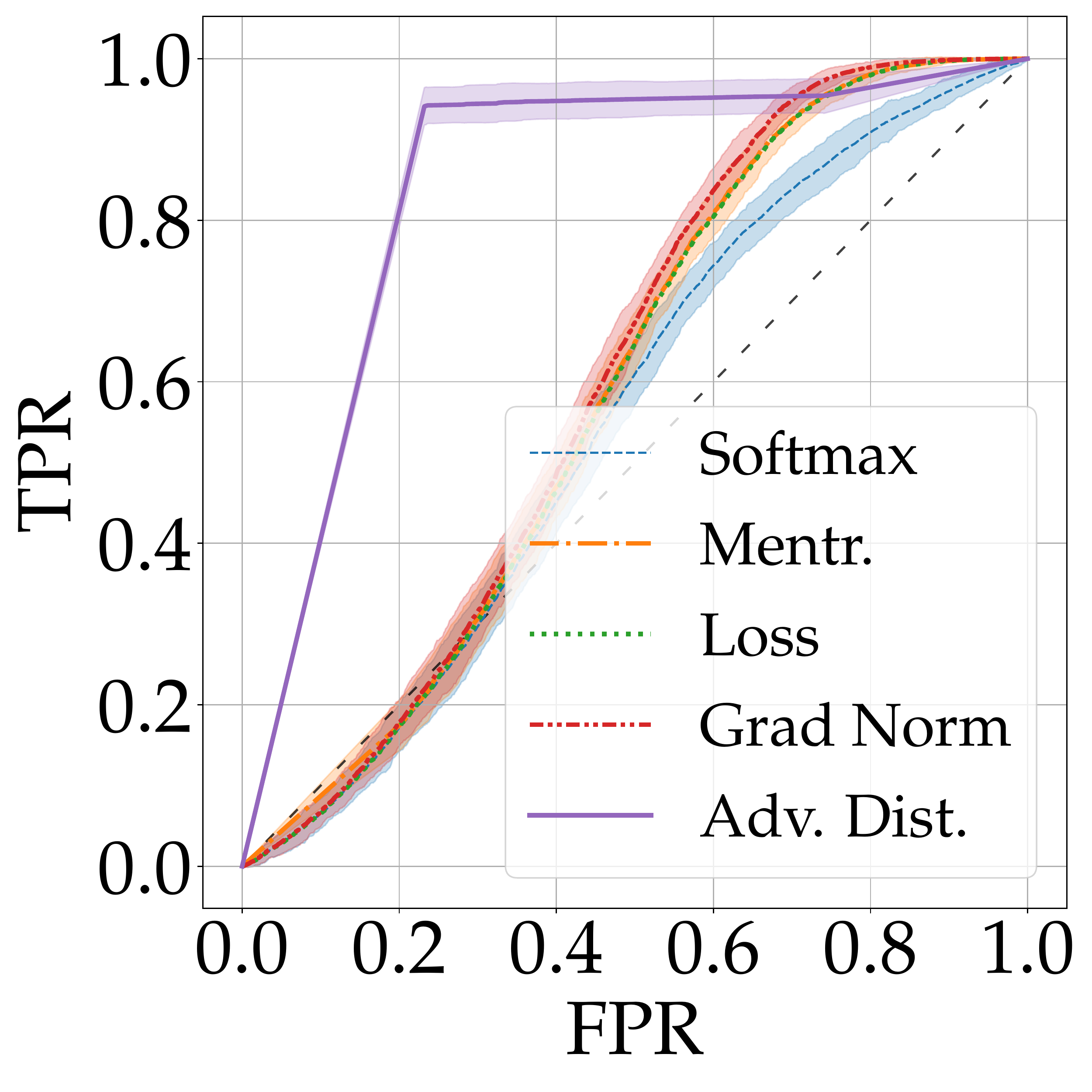}
            \caption[]%
            {{\small ResNet}}    
            \label{fig:ResNet_}
        \end{subfigure}
        %\hfill
        \begin{subfigure}[b]{0.24\textwidth}   
            \centering 
 \includegraphics[width=\textwidth]{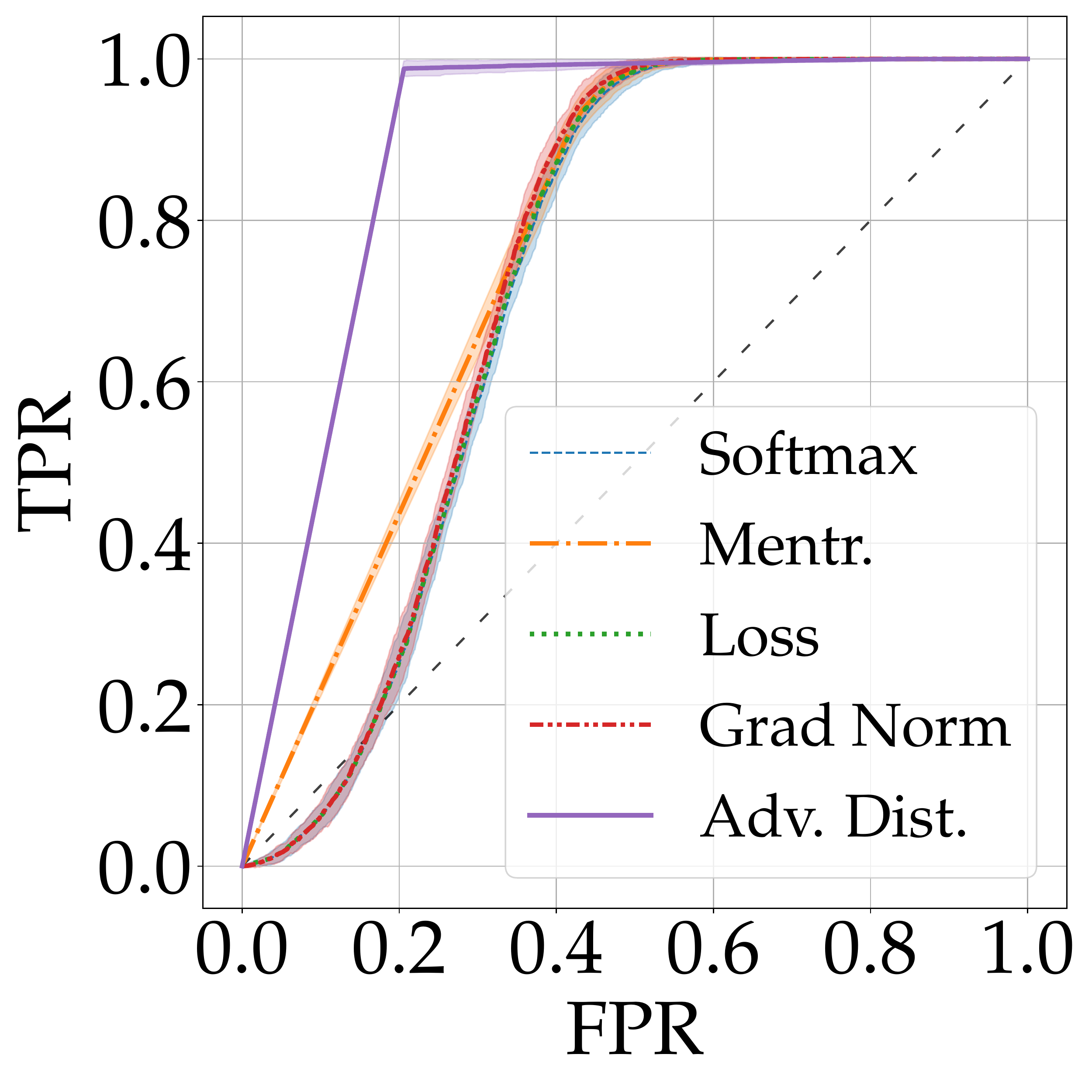}
            \caption[]%
            {{\small ResNext}}    
            \label{fig:Resnext_}
        \end{subfigure}
        \caption[ The average and standard deviation of critical parameters ]
        {\small ROC curves averaged over $20$ iterations on AlexNet (\ref{fig:AlexNet_}), 
        DenseNet (\ref{fig:DenseNet_}),
        ResNet (\ref{fig:ResNet_}),
        ResNext (\ref{fig:Resnext_}). The confidence interval correspond to $10$ time the standard deviation.}
        \label{fig:rocs_}
\end{figure*}
\Cref{fig:rocs_} is a copy of \cref{fig:rocs}, except it includes the confidence intervals for the TPR. The curves in these figures are computed using the following process: A grid of FPR values is fixed. For each run of the experiment, we compute a ROC curve and use it to interpolate the TPR values that correspond to the fixed FPR values. The TPR values are averaged over different runs of the experiment. 

The confidence intervals presented in \cref{fig:rocs_} correspond to $10$ times the standard deviation. We chose to multiply the standard deviation in order to make the confidence intervals visible in the figure. Indeed, in our experimental setting as the train and test sets have large sizes, the ROC curves of MIA strategies remain close to the average ROC curves over several shots of train and test sets. 

\Cref{tab:AUROC1} contains the results for additional attack strategies that were not included in the main body of the paper. 

Among the strategies that can be implemented as a binary decision test, we include the Grad $x$ Norm strategy, and the adversarial distance strategy with other different norms. In the Grad $x$ Norm strategy, the gradient of the loss function of the target model with respect to the input image is computed. Then, the $\ell_2$ norm of the gradient is computed and used for the binary decision test. The Grad $w$ Norm is equivalent to the Grad Norm strategy presented in the main body of the paper. In the main body of the paper, we present the results for the adversarial distance strategy that uses the $\ell_\infty$ norm to measure the distance between samples. Here we include the results for the $\ell_1$ and $\ell_2$ norms.

\Cref{tab:AUROC1} also shows the results for two additional models that require training. Namely, the ensemble attacker explained in \cref{sec:ensemble} and referred to as ML attacker in the table, and the Logits attacker from \cite{Rezaei}. The Logits attacker is similar to the intermediate outputs attacker explained in the main body of the paper, but only utilizes the outputs of the last layer.

\begin{table*}[t]
    \centering
    \resizebox{\textwidth}{!}{
    \begin{tabular}{l|c|c|c|c|c|c|c|c}
    Attack & \multicolumn{2}{c|}{AlexNet} & \multicolumn{2}{c|}{ResNet} & \multicolumn{2}{c|}{ResNext} & \multicolumn{2}{c}{DenseNet} \\
    \cline{2-9}
    Strategy & AUROC & Accuracy & AUROC & Accuracy & AUROC & Accuracy & AUROC & Accuracy  \\
    \hline
    Softmax & $68.00\pm0.16$ & $65.34\pm0.14$ & $55.45\pm0.15$ & $57.40\pm0.13$ & $72.37\pm0.07$ & $74.84\pm0.11$ & $70.52\pm0.09$ & $72.11\pm0.10$ \\
    Mentr. \cite{Song2021} & $77.11\pm0.10$ & $74.16\pm0.11$ & $59.10\pm0.13$ & $61.39\pm0.11$ & $76.87\pm0.08$ & $75.28\pm0.11$ & $74.21\pm0.10$ & $72.69\pm0.10$ \\
    Loss & $76.69\pm0.10$ & $74.14\pm0.13$ & $58.66\pm0.13$ & $61.29\pm0.11$ & $72.57\pm0.07$ & $75.17\pm0.11$ & $70.85\pm0.09$ & $72.61\pm0.10$ \\
    Grad $w$ Norm & $76.58\pm0.10$ & $74.19\pm0.12$ & $59.93\pm0.13$ & $62.56\pm0.09$ & $73.06\pm0.07$ & $75.74\pm0.11$ & $71.30\pm0.09$ & $73.81\pm0.09$ \\
    Grad $x$ Norm & $75.20\pm0.14$ & $73.12\pm0.12$ & $59.64\pm0.14$ & $62.17\pm0.11$ & $72.92\pm0.07$ & $75.62\pm0.10$ & $71.15\pm0.08$ & $73.00\pm0.09$ \\
    Adv. Dist. $\|\cdot\|_\infty$ & $84.35\pm0.13$ & $85.12\pm0.18$ & $84.53\pm0.16$ & $85.45\pm0.11$ & $89.24\pm0.03$ & $89.10\pm0.05$ & $82.76\pm0.03$ & $82.63\pm0.05$ \\
    Adv. Dist. $\|\cdot\|_2$ & $76.89\pm0.16$ & $74.03\pm0.15$ & $70.86\pm0.19$ & $67.71\pm0.18$ & $72.47\pm0.19$ & $67.68\pm0.16$ & $68.00\pm0.21$ & $62.65\pm0.20$ \\
    Adv. Dist. $\|\cdot\|_1$ & $73.66\pm0.12$ & $74.03\pm0.10$ & $64.94\pm0.16$ & $64.51\pm0.13$ & $70.98\pm0.19$ & $66.95\pm0.15$ & $59.39\pm0.17$ & $59.27\pm0.04$ \\
    \hline
    {\MLattacker}* & $90.84\pm0.13$ & $85.48\pm0.67$ & $89.31\pm1.04$ & $85.07\pm0.47$ & $92.30\pm0.19$ & $92.17\pm0.15$ & $87.86\pm0.22$ & $87.46\pm0.20$ \\
    Grad $w$* \cite{Rezaei} & $78.76\pm0.30$ & $74.32\pm0.28$ & $61.98\pm0.38$ & $62.72\pm0.27$ & $77.80\pm0.30$ & $73.47\pm0.57$ & $73.12\pm1.42$ & $72.59\pm0.55$ \\
    Grad $x$* \cite{Rezaei} & $77.20\pm0.26$ & $73.43\pm0.26$ & $68.48\pm0.27$ & $63.58\pm0.22$ & $77.54\pm0.61$ & $73.47\pm0.57$ & $75.81\pm0.43$ & $71.81\pm0.40$ \\
    Int. Outs* \cite{Rezaei} & $57.92\pm0.50$ & $56.36\pm0.41$ & $96.59\pm0.29$ & $91.57\pm0.43$ & $93.62\pm0.39$ & $86.38\pm0.37$ & $99.17\pm0.10$ & $97.68\pm0.14$ \\
    Logits* \cite{Rezaei} & $58.19\pm0.57$ & $56.35\pm0.52$ & $88.96\pm0.44$ & $81.38\pm0.57$ & $84.89\pm0.37$ & $77.24\pm0.33$ & $67.64\pm0.57$ & $63.19\pm0.46$ \\
    WB* \cite{NasrShokri} & $80.33\pm1.21$ & $74.03\pm0.71$ & $87.51\pm0.41$ & $79.73\pm0.30$ & $84.52\pm1.95$ & $76.46\pm1.82$ & $79.38\pm1.16$ & $71.92\pm0.97$ \\
    \end{tabular}
    }
    \caption{Comparison of different MIA Techniques. The Accuracy($\%$) and AUROC score ($\%$) on a balanced evaluation set are reported. $10k$ are uniformly selected from the training set (members) and the whole $10k$ samples from the testing set are selected (non-members). All the data selected is used for evaluation. Techniques with a (*) require training. In this case, only $60\%$ of the data is used for evaluation and rest is used for training.}
    \label{tab:AUROC1}
\end{table*}

\section{Additional Results for the ML Attacker}
\label{sec:additionalresults}

We also study how the amount of side information influences the performance of the attacker. \Cref{tab:MLattacker} reports the AUROC score of the {\MLattacker} against the pre-trained target models for different amounts of side information. The side information is always composed by $50\%$ \textit{in-training} samples and $50\%$ \textit{out-of-training} samples, and the total is indicated in the table. Remark that the performance of the {\MLattacker} might improve with the increase in the size of its training set; however, in some cases a small training set ($1000$ samples) is enough to obtain an effective attack model.~
\begin{table}[h]
    \centering
    \resizebox{.475\textwidth}{!}{%
    \begin{tabular}{l|c|c|c|c}
    Target & \multicolumn{4}{c}{Training set size} \\
    \cline{2-5}
    Model & $1000$ & $2000$ & $4000$ & $8000$ \\
    \hline
    AlexNet & $86.68\pm1.95$ & $88.25\pm1.63$ & $89.87\pm0.73$ & $90.84\pm0.52$  \\
    ResNet & $87.15\pm1.05$ & $88.34\pm0.68$ &$89.26\pm0.55$ & $89.31\pm1.04$  \\
    ResNext & $92.24\pm0.22$ & $92.30\pm0.28$ &$92.31\pm0.19$ & $92.30\pm0.19$  \\
    DenseNet & $86.20\pm5.51$ & $87.42\pm1.45$ &$87.71\pm0.56$ & $87.58\pm0.22$  \\
    \end{tabular}%
    }
    \caption{Influence of training set size on performance for the {\MLattacker}. The AUROC ($\%$) for a balanced evaluation set is reported. Half of the training samples for the attacker are uniformly selected from the original training set and the other half from the test set. Then, $6k$ samples from the training set and $6k$ samples from the test set are uniformly selected for evaluation.}
    \label{tab:MLattacker}
\end{table}

\end{document}